\documentclass[fleqn,10pt]{wlscirep}
\usepackage[utf8]{inputenc}
\usepackage[T1]{fontenc}
\usepackage{subcaption}
\usepackage{tikz}
\usepackage{xcolor}

\usepackage{url}

\usetikzlibrary{shapes,arrows}
\usetikzlibrary{shapes.multipart}
 
\definecolor{filtering_step_color}{HTML}{FFEFD5}
\definecolor{create_task_color}{HTML}{FFEFD5}
\definecolor{exclude_color}{HTML}{FFE4E1}
\definecolor{benchmark_task_color}{HTML}{66CDAA}
\definecolor{cohort_color}{HTML}{E6E6FA}

\tikzset{
    use bounding box relative coordinates/.style={
        shift={(current bounding box.south west)},
        x={(current bounding box.south east)},
        y={(current bounding box.north west)}
    },
}

\definecolor{darkgreen}{HTML}{2E8B57}

\usetikzlibrary{arrows,decorations.markings}
\usetikzlibrary{decorations.pathreplacing}

\title{Multitask learning and benchmarking with \\clinical time series data}

\author[1]{Hrayr Harutyunyan}
\author[2,3]{Hrant Khachatrian}
\author[1]{David C. Kale}
\author[1]{Greg Ver Steeg}
\author[1]{Aram Galstyan}
\affil[1]{USC Information Sciences Institute, Marina del Rey, California 90292, United States of America}
\affil[2]{YerevaNN, Yerevan, 0025, Armenia}
\affil[3]{Yerevan State University, Yerevan, 0025, Armenia}

\affil[*]{hrant@yerevann.com}

\begin{abstract}
Health care is one of the most exciting frontiers in data mining and machine learning.
Successful adoption of electronic health records (EHRs) created an explosion in digital clinical data available for analysis, but progress in machine learning for healthcare research has been difficult to measure because of the absence of publicly available benchmark data sets.
To address this problem, we propose four clinical prediction benchmarks using data derived from the publicly available Medical Information Mart for Intensive Care (MIMIC-III) database.
These tasks cover a range of clinical problems including modeling risk of mortality, forecasting length of stay, detecting physiologic decline, and phenotype classification.
We propose strong linear and neural baselines for all four tasks and evaluate the effect of deep supervision, multitask training and data-specific architectural modifications on the performance of neural models.
\end{abstract}

\begin{document}

\flushbottom
\maketitle
\thispagestyle{empty}

\section*{Introduction}
\label{sec:introduction}

In the United States alone, each year over 30 million patients visit hospitals~\cite{healthcare2014introduction}, 83\% of which use an electronic health record (EHR) system~\cite{henryadoption}.
This trove of digital clinical data presents a significant opportunity for data mining and machine learning researchers to
solve pressing health care problems, such as
early triage and risk assessment, prediction of physiologic decompensation, identification of high cost patients, and characterization of complex, multi-system diseases~\cite{bates2014big,zimmerman2006acute,williams2012national,dahl2012high,saria2015subtyping}.
These problems are not new (the word \textit{triage}, dates back to at least World War I and possibly earlier~\cite{iserson2007triage}, while the Apgar risk score was first published in 1952~\cite{apgar1952proposal}), but the success of machine learning~\cite{ferrucci2013watson,silver2016mastering} and growing availability of clinical data have sparked widespread interest.

While there has been a steady growth in machine learning research for health care, several obstacles have slowed progress in harnessing digital health data. The main challenge is
the absence of widely accepted benchmarks to evaluate competing models.
Such benchmarks accelerate progress in machine learning by focusing the community
and facilitating reproducibility and competition.
For example, the winning error rate in the ImageNet Large Scale Visual Recognition Challenge (ILSVRC) plummeted
an order of magnitude from 2010 (0.2819) to 2016 (0.02991).
In contrast, practical progress in clinical machine learning has been difficult to measure due to variability in data sets and task definitions~\cite{caballero2015dynamically,ghassemi2015multivariate,luo2016predicting,lee2017customization,johnson2017mlhc}.
Public benchmarks also lower the barrier to entry by enabling new researchers to start without having to negotiate data access or recruit expert collaborators. 

Additionally, most of the researchers develop new methods for one clinical prediction task at a time (e.g., mortality prediction~\cite{caballero2015dynamically}
or condition monitoring~\cite{quinn2009factorial}).
This approach is detached from the realities of clinical decision making, in which all the above tasks are often performed simultaneously by clinical staff~\cite{Laxmisan2007801}. Perhaps more importantly, there is accumulating evidence that those prediction tasks are interrelated. For instance, the highest risk and highest cost patients are often those with complex comorbidities~\cite{horn1991relationship} while decompensating patients have a higher risk for poor outcomes~\cite{williams2012national}.

In this paper, we take a comprehensive approach to addressing the above challenges.
We propose a public benchmark suite that includes four different clinical prediction tasks inspired by the opportunities for ``big clinical data'' discussed in Bates et al.~\cite{bates2014big}: in-hospital mortality, physiologic decompensation, length of stay (LOS), and phenotype classification.
Derived from the publicly available Medical Information Mart for Intensive Care (MIMIC-III) database~\cite{johnson2016mimic,mimicdata}, 
our benchmark contains rich multivariate time series
from over 40,000 intensive care unit (ICU) stays as well as labels for four tasks spanning
a range of classic machine learning problems from multilabel time series classification to regression with skewed responses.
These data are suitable for research on topics as diverse as non-random missing data and time series analysis.


This setup of benchmarks allows to formulate a \textit{heterogeneous multitask learning} problem that involves jointly learning all four prediction tasks simultaneously.
These tasks vary in not only output type but also temporal structure: LOS involves a regression at each time step, while in-hospital mortality risk is predicted once early in admission.
Their heterogeneous nature requires a modeling solution that can not only handle sequence data but also model correlations between tasks distributed in time.
We demonstrate that carefully designed recurrent neural networks are able to exploit these correlations to improve the performance for several tasks. 

Our code is already available online \cite{mimicbenchmarkrepo}, so that anyone with access to MIMIC-III can build our benchmarks and reproduce our experiments sidestepping difficulties of preprocessing of clinical data.

\subsection*{Related Work}
\label{sec:relatedwork}

There is an extensive body of research on clinical predictions using deep learning, and we will attempt to highlight only the most representative or relevant work since a full treatment is not possible.

Feedforward neural networks nearly always outperform logistic regression and severity of illness scores
in modeling mortality risk among hospitalized patients~\cite{caruana1996using,clermont2001predicting,celi2012database}. 
Recently, it was shown that novel neural architectures (including ones based on LSTM) perform well for predicting inpatient mortality, 30-day unplanned readmission, long length-of-stay (binary classification) and diagnoses on general EHR data (not limited to ICU)~\cite{GoogleBrainEHR2018}. The experiments were done on several private datasets. 

There is a great deal of early research that uses neural networks to predict LOS in hospitalized patients~\cite{grigsby1994simulated,mobley1995artificial}.
However, rather than regression, much of this work formulates the task as binary classification aimed at identifying patients at risk for 
long stays~\cite{buchman1994comparison}. 
Recently, novel deep learning architectures have been proposed for
survival analysis~\cite{yousefi2016learning,ranganath2016deep},
a similar time-to-event regression task with right censoring.

Phenotyping has been a popular application for deep learning researchers in recent years, though model architecture and problem definition vary widely.
Feedforward networks~\cite{lasko2013plosone,che2015deep}, LSTM networks~\cite{choi2015doctor} and temporal convolutional networks~\cite{razavian2016multi} have been used to predict diagnostic codes from clinical time series.
In 2016, it was first shown that recurrent neural networks could classify dozens of acute care diagnoses in variable length clinical time series~\cite{lipton2016learning}.

Multitask learning has its roots in clinical prediction~\cite{caruana1996using}. 
Several authors formulated phenotyping as multi-label classification, using neural networks to implicitly capture comorbidities in hidden layers~\cite{lipton2016learning,razavian2016multi}.
Others attempted to jointly solve multiple 
related clinical tasks, including predicting mortality and length of stay~\cite{ngufor2015multi}.
However, none of this work addressed problem settings where sequential or temporal structure varies across tasks.
The closest work in spirit to ours is a paper by Collobert and Weston~\cite{collobert2008unified} where a single convolutional network is used to perform a variety of natural language tasks (part-of-speech tagging, named entity recognition, and language modeling) with diverse sequential structure.





Earlier version of this work has been available online for two years (arXiv:1703.07771v1). The current version adds more detailed description of the dataset generation process, improves the neural baselines and adds more discussion on the results. Since the release of the preliminary version of the benchmark codebase, several teams used our dataset generation pipeline (fully or partially). In particular, the pipeline was used for in-hospital mortality prediction~\cite{TCS-first,TCS-transfer,amazon-NER-2018,Clinical-SequenceTransformerNetworks2018,MissingDataRepresentations2019,AdverseEventForecastingRL2018}, decompensation prediction~\cite{RAIM2018}, length-of-stay prediction~\cite{RNN-GP2018chung,RAIM2018,MissingDataRepresentations2019}, phenotyping~\cite{taha,TCS-first,TCS-transfer} and readmission prediction~\cite{TAMUreadmissions}. Additionally, attention-based RNNs were applied for all our benchmark tasks~\cite{attendanddiagnose}.

In a parallel work another set of benchmark tasks based on MIMIC-III was introduced that includes multiple versions of in-hospital mortality predictions, length-of-stay and ICD-9 code group predictions, but does not include decompensation prediction~\cite{purushotham2017benchmark}.
The most critical difference is that in all their prediction tasks the input is either the data of the first 24 or 48 hours, while we do length of stay and decompensation prediction at each hour of the stay, and do phenotyping based on the data of the entire stay.
We frame the length-of-stay prediction as a classification problem and use the Cohen's kappa score as its metric, while they frame it as a regression problem and use the mean squared error as its metric. 
The metric they use is less indicative of performance given that the distribution of length of stay has a heavy tail.
In ICD-9 code group prediction, we have 25 code groups as opposed to their 20 groups.
There are many differences in the data processing and feature selection as well.
For example, we exclude all
ICU stays where the patient is younger than 18, while they exclude patients younger than 15.
Moreover, they consider only the first admission of a patient, while we consider all admissions.
They have benchmarks for three different features sets: A, B, and C, while we have only one set of features, which roughly corresponds to their feature set A.
The set of baselines is also different. While our work has more LSTM-based baselines, the parallel work has more baselines with traditional machine learning techniques.

\section*{Results}
\label{sec:results}

\begin{figure}[t!]
    \centering
    \includegraphics[width=0.99\textwidth]{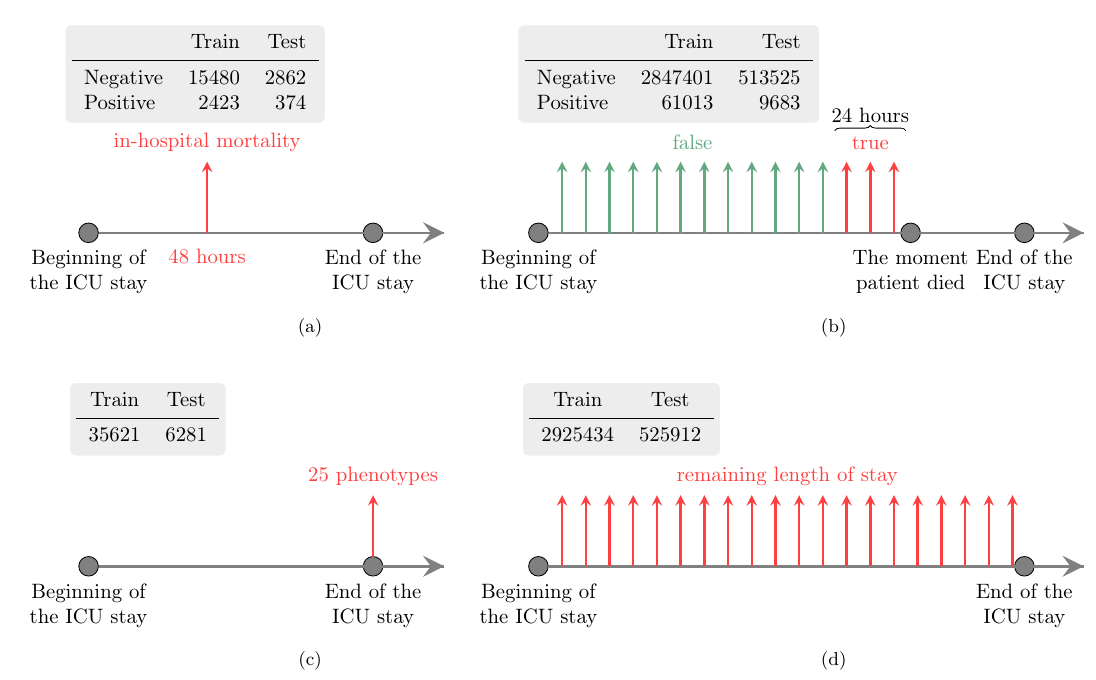}
    \caption{Summaries of the four benchmark tasks. Each subfigure consists of two parts. The table lists number of prediction instances for the corresponding task. The timeline shows when the predictions are done. Note that in the decompensation and length-of-stay prediction tasks predictions are done hourly, and each vertical arrow corresponds to one prediction instance. (a) In-hospital mortality. (b) Decompensation. (c) Phenotyping. (d) Length of stay.}
    \label{fig:tasks}
\end{figure}

We compile a subset of the MIMIC-III database containing more than 31 million clinical events that correspond to 17 clinical variables listed in the first column of Table \ref{tab:variables}. These events cover 42276 ICU stays of 33798 unique patients. 
We define four benchmark tasks on this subset.
\begin{enumerate}
    \item In-hospital mortality prediction -- predicting in-hospital mortality based on the first 48 hours of an ICU stay. This is a binary classification task with area under the receiver operating characteristic (AUC-ROC) being the main metric.
    \item Decompensation prediction -- predicting whether the patient's health will rapidly deteriorate in the next 24 hours. The goal of this task is to replace early warning scores currently used in the hospitals. Due to the lack of gold standard for evaluating the early warning scores, we follow earlier work \cite{williams2012national} and define our task as mortality prediction in the next 24 hours at each hour of an ICU stay. It is important to note that this definition deviates from the core meaning of decompensation, and the task becomes similar to the first one. On the other hand, we believe this is the closest proxy task for decompensation prediction for which one can obtain precise labels from MIMIC-III database. Each instance of this task is a binary classification instance. Likewise in-hospital mortality prediction, the main metric is AUC-ROC.
    \item Length-of-stay prediction -- predicting remaining time spent in ICU at each hour of stay. Accurate prediction of the remaining length-of-stay is important for scheduling and hospital resource management. We frame this as a classification problem with 10 classes/buckets (one for ICU stays shorter than a day, seven day-long buckets for each day of the first week, one for stays of over one week but less than two, and one for stays of over two weeks). The main metric for this task is Cohen's linear weighted kappa score.
    \item Phenotype classification --  classifying which of 25 acute care conditions (described in Table \ref{tab:phenotypes}) are present in a given patient ICU stay record. This problem is a multilabel classification problem with macro-averaged AUC-ROC being the main metric.
\end{enumerate}
Additionally, a multitask version of the four tasks is defined.
The tasks are summarized in the Figure \ref{fig:tasks}.

\begin{figure}[tbh!]
\centering
\includegraphics[width=0.99\textwidth]{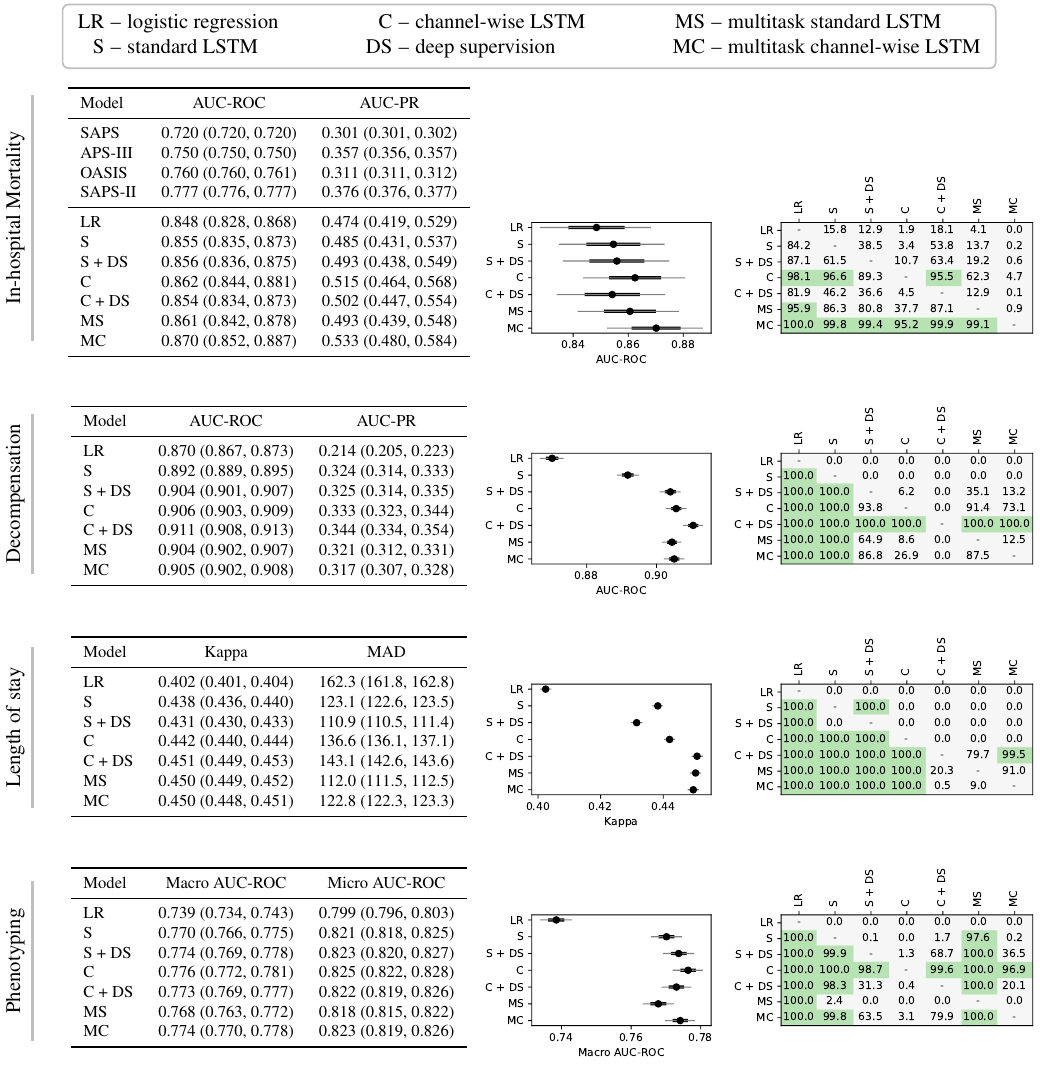}
\caption{Results for in-hospital mortality, decompensation, length-of-stay, and phenotype prediction tasks. A subfigure corresponding to a benchmark task has three parts. The first part is a table that lists the values of the metrics for all models along with 95\% confidence intervals obtained by resampling the test set $K$ times with replacement ($K=10000$ for in-hospital mortality and phenotype prediction task, while for decompensation and length-of-stay prediction tasks $K=1000$). For all metrics except MAD, larger values are better. 
The second part visualizes the confidence intervals for the main metric of the corresponding task. The black circle corresponds to the mean value of $K$ iterations.
The thick black line shows standard deviation and narrow grey line shows 95\% confidence interval.
The third part shows the significance of the difference between the models. 
We count the number of resampled tests sets on which the $i$-th model performed better than the $j$-th model (denoted by $c_{i,j}$).
The cell at the $i$-th row and the $j$-th column of the table shows the percentage of $c_{i,j}$ in $K$. We say that the $i$-th model is significantly better than the $j$-th model if $c_{i,j}/K > 0.95$ and highlight the corresponding cell of the table.
}
\label{fig:results-summary}
\end{figure}

We develop linear regression models and multiple neural architectures for the benchmark tasks. We perform experiments with a basic LSTM-based neural network (standard LSTM) and introduce a modification of it (channel-wise LSTM). Additionally, we test both types of LSTMs with deep supervision and multitask training. We perform a hyperparameter search to select the best performing models and evaluate them on the test sets of the corresponding tasks. 
By doing bootstrapping on the test set we also report 95\% confidence intervals for the models and test the statistical significance of the differences between models. The results for each of the mortality, decompensation, LOS, and phenotyping tasks are reported in Figure \ref{fig:results-summary}.


We first note that LSTM-based models outperformed linear models by substantial margins across all metrics on every task. The difference is significant in every case except three out of six LSTM models for in-hospital mortality.
This is consistent with previous research comparing neural networks to linear models for mortality prediction~\cite{clermont2001predicting}, and phenotyping~\cite{lipton2016learning} but it is nonetheless noteworthy because questions still remain about the potential effectiveness of deep learning for health data, especially given the often modest size of the data relative to their complexity.
Our results provide further evidence that complex architectures can be effectively trained on non-Internet scale health data and that while challenges like overfitting persist, they can be mitigated with careful regularization schemes, including dropout and multitask learning.

The experiments show that channel-wise LSTMs and multitask training act as regularizers for almost all tasks. Channel-wise LSTMs perform significantly better than standard LSTMs for all four tasks, while multitasking helps for all tasks except phenotyping (the difference is significant for decompensation and length-of-stay prediction tasks). We hypothesize that this is because phenotype classification is itself a multitask problem and already benefits from regularization by sharing LSTM layers across the 25 different phenotypes. The addition of further tasks with loss weighting may limit the multitask LSTM's ability to effectively learn to recognize individual phenotypes. Note that the hyperparameter search for multitask models did not include zero coefficients for any of the four tasks. That is why the best multitask models sometimes perform worse than single-task models..
The combination of the channel-wise layer and multitasking is also useful. Multitask versions of channel-wise LSTMs perform significantly better than the corresponding single-task versions for in-hospital mortality prediction and phenotyping tasks.

Deep supervision with replicated targets did not help for in-hospital mortality prediction. For phenotyping, it helped for the standard LSTM model (as discovered in an earlier work~\cite{lipton2016learning}), but did not help for channel-wise models. On the other hand, we see significant improvements from deep supervision for decompensation and length-of-stay prediction tasks (except for the Standard LSTM model for length-of-stay prediction). For both these tasks the winner models are channel-wise LSTMs with deep supervision. For decompensation, the winner is significantly better than all other models and for LOS the winner is significantly better than all others except the runner-up model, which is a multitask standard LSTM.

\section*{Discussion}\label{sec:discussion}

In this paper we proposed four standardized benchmarks for machine learning researchers interested in clinical data problems, including in-hospital mortality, decompensation, length-of-stay, and phenotype prediction.
Our benchmark data set is similar to other MIMIC-III patient cohorts described in machine learning publications but makes use of a larger number of patients and is immediately accessible to other researchers who wish to replicate our experiments or build upon our work.
We also described several strong baselines for our benchmarks. We have shown that LSTM-based models significantly outperform linear models, although we expect to see better performing linear models by using more complex feature engineering. We have demonstrated the advantages of using channel-wise LSTMs and learning to predict multiple tasks using a single neural model.
Our results indicate that the phenotyping and length-of-stay prediction tasks are more challenging and require larger model architectures than mortality and decompensation prediction tasks.
Even small LSTM models easily overfit the latter two problems.

We note that since the data in MIMIC-III is generated within a single EHR system, it might contain systematic biases. It is an interesting future study to explore how models trained on these benchmarks generalize to other clinical datasets. 

\subsection*{In-hospital mortality prediction}

\begin{figure}[tb!]
\centering
\includegraphics[width=0.99\textwidth]{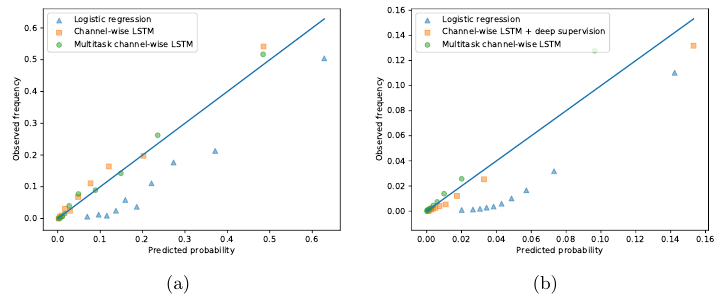}
\caption{Calibration of in-hospital mortality and decompensation prediction by the best linear, non-multitask and multitask LSTM-based  models. The plots show predicted probability computed by creating decile bins of predictions and then taking the
mean value within each bin vs. actual probability (the rate of mortality within each bin).
(a) In-hospital mortality. (b) Decompensation.}
\label{fig:calibration}
\end{figure}

\begin{figure}[tb!]
\centering
\includegraphics[width=0.99\textwidth]{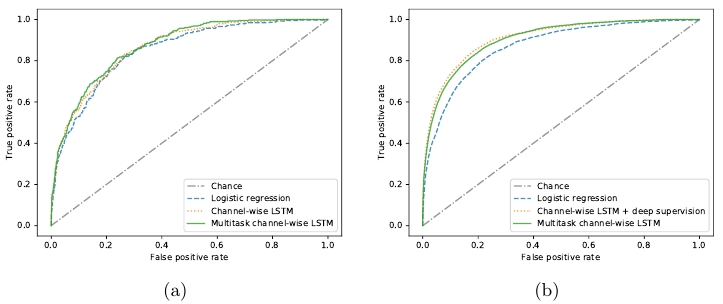}
\caption{Receiver operating characteristic curves for the best linear, non-multitask and multitask LSTM-based models. (a) In-hospital mortality. (b) Decompensation}
\label{fig:roc_curves}
\end{figure}

For risk-related tasks like mortality and decompensation, we are also interested how reliable the probabilities estimated by our predictive models are.
This is known as \textit{calibration} and is a common method for evaluating predictive models in the clinical research literature.
In a well calibrated model, 10\% all patients who receive a predicted 0.1 probability of decompensation do in fact decompensate. 
We included no formal measure of calibration in our benchmark evaluations, but we informally visualize calibration for mortality and decompensation predictions using reliability plots.
These are scatter plots of predicted probability vs. actual probability.
Better calibrated predictions will fall closer to the diagonal.
Figure \ref{fig:calibration} (a) visualizes calibration of several in-hospital mortality prediction baselines.
We see that the LSTM-based models look reasonably calibrated, while the logistic regression baseline consistently overestimates the actual probability of mortality.
We would like to note that the logistic regression model can be successfully calibrated using Platt scaling or isotonic regression.

\begin{figure}[t]
\centering
\includegraphics[width=0.99\textwidth]{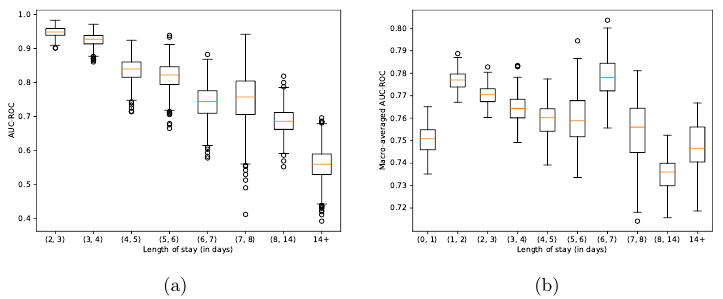}
\caption{In-hospital mortality and phenotype prediction performance vs. length-of-stay. The plots show the performance of the best non-multitask prediction baselines on the test data of different length-of-stay buckets.
The confidence intervals and standard deviations are estimated with bootstrapping on the data of each bucket. (a) In-hospital mortality. (b) Phenotype.}
\label{fig:ihm_pheno_vs_los}
\end{figure}

To provide more insights on the results of this task we present the receiver operating characteristic (ROC) curves corresponding to the best logistic, non-multitask and multitask LSTM-based in-hospital mortality prediction models in Figure \ref{fig:roc_curves} (a).
Additionally, with Figure \ref{fig:ihm_pheno_vs_los} (a) we demonstrate, perhaps unsurprisingly, that the performance of the best non-multitask baseline (channel-wise LSTM) degrades as length-of-stay increases.
Finally, in the in-hospital mortality part of the Figure \ref{fig:results-summary} we compare our baselines with the traditional scores like SAPS~\cite{le1984simplified}. 
Note that these scores are computed on the first 24 hours of the stays.

Out of the four tasks, in-hospital mortality prediction task is the only one for which most of the differences between different baselines are not significant.
The reason behind this is the relatively small size of the test set (about twice smaller than that of phenotype prediction task and much more smaller than those of length-of-stay and decompensation prediction tasks).

\subsection*{Decompensation prediction}


\begin{figure}[t]
\begin{center}
\includegraphics[width=0.5\textwidth]{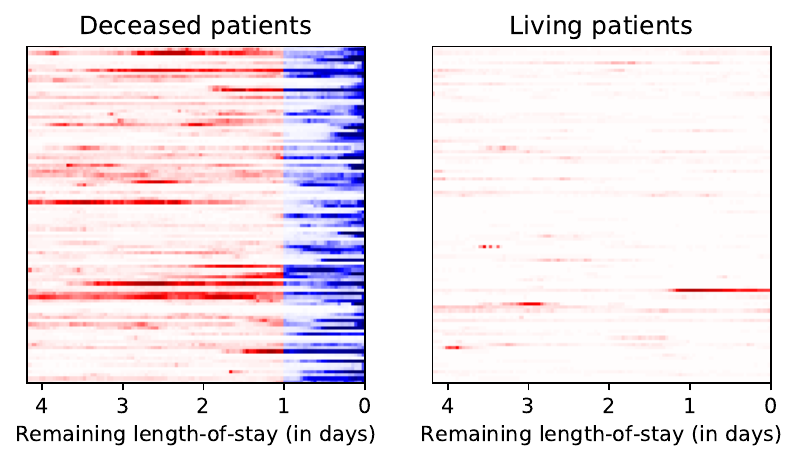}
\end{center}
\caption{Prediction of channel-wise LSTM baseline with deep supervision for decompensation prediction over time. Each row shows the last 100 hours of a single ICU stay. Darker colors mean high probability predicted by the model. Red and blue colors indicate the ground-truth label is negative and positive, respectively. Ideally, the right image should be all white, and the left image should be all white except the right-most 24 hours, which should be all dark blue.}
\label{fig:decomp_prediction}
\end{figure}

Figure \ref{fig:calibration} (b) visualizes calibration of several decompensation prediction baselines.
Likewise the case of in-hospital mortality prediction task, we see that the LSTM-based models are better calibrated than the logistic regression model.
Again, the logistic regression model can be reasonably calibrated using Platt scaling or isotonic regression.
Compared to in-hospital mortality baselines, we see that decompensation baselines are worse calibrated.
This behavior is expected since the decompensation prediction has more severe class imbalance.

To provide further insights on the results we present the ROC curves corresponding to the best logistic, non-multitask and multitask LSTM-based decompensation prediction models in Figure \ref{fig:roc_curves} (b).
Additionally, to understand better what the best decompensation prediction model (channel-wise LSTM with deep supervision) does, we visualize its predictions over the time in Figure \ref{fig:decomp_prediction}. The left part of the figure shows randomly chosen 100 patients from the test set that died in ICU. The right part shows another 100 patients randomly chosen from the test set. Every row shows the predictions for the last 100 hours of a single ICU stay. Darker colors indicate higher predicted probability of death in the upcoming 24 hours. Red and blue colors indicate ground truth labels (blue is positive mortality). The right part of the figure shows that for living patients the model rarely produces false positives. The left part shows that in many cases the model predicts mortality days before the actual time of death. On the other hand, there are many cases when the mortality is predicted only in the last few hours, and in a few cases the model doesn't give high probability even at the last hour. This figure shows that even a model with 0.91 AUC-ROC can make trivial mistakes and there is a lot of room for improvement. 

\subsection*{Length-of-stay prediction}

For length-of-stay prediction task we also tried regression models that directly predict the number of days. These models consistently performed worse than classification models in terms of kappa score, but had better mean absolute difference (MAD), as shown in Table \ref{tab:los_reg}. Note that the regression models for LOS prediction were trained with mean squared error loss function. Therefore, the MAD scores they get are suboptimal and can be improved by directly training to minimize the MAD score.
In general, our results for LOS forecasting are the worst among the four tasks.
Our intuition is that this due in part to the intrinsic difficulty of the task, especially distinguishing between stays of, e.g., 3 and 4 days.

To investigate this intuition further, we considered a task formulation similar to the one described in Rajkomar et al.~\cite{GoogleBrainEHR2018} where the goal was to predict whether a patient would have an extended LOS (longer than seven days) from only the first 24 hours of data. 
In order to evaluate our models in a similar manner, we summed the predicted probabilities from our multiclass LOS model for all buckets corresponding to seven days or longer LOS.
For our best LOS model, this yielded an AUC-ROC of 0.84 for predicting extended LOS at 24 hours after admission.
This is comparable to the results from Rajkomar et al. who reported AUC-ROCs of 0.86 and 0.85 on two larger private datasets using an ensemble of several neural architectures.
This is especially noteworthy since our models were not trained to solve this particular problem and suggests that the extended LOS problem is more tractable than the regression or multiclass versions.
Nonetheless, solving the more difficult fine-grained LOS problem remains an important goal for clinical machine learning researchers.


\begin{table}[b]
\centering
\caption{Results for length of stay prediction task (regression). Contrary to mean absolute deviation (MAD), larger kappa is better.}
\label{tab:los_reg}
\begin{tabular}{lcc}
\toprule
Model      &             Kappa              &              MAD               \\
\midrule
Linear regression       &    0.336 (0.335, 0.338)     &    116.4 (115.8, 117.0)     \\
Standard LSTM          &    0.433 (0.432, 0.435)     &      94.7 (94.2, 95.1)      \\
Standard LSTM + deep supervision     &    0.413 (0.411, 0.414)     &      94.5 (94.1, 95.0)      \\
Channel-wise LSTM         &    0.424 (0.422, 0.425)     &      94.3 (93.9, 94.8)      \\
Channel-wise LSTM + deep supervision     &    0.426 (0.424, 0.428)     &      94.0 (93.6, 94.4)      \\
\bottomrule
\end{tabular}
\end{table}

\subsection*{Phenotyping}

Phenotyping is actually a combination of 25 separate binary classification tasks and the performance of the models vary across different tasks.
Table \ref{tab:phenotypes} shows the per-phenotype ROC-AUC for the best phenotype baseline (channel-wise LSTM).
We observe that AUC-ROC scores on the individual diseases vary widely from 0.6834 (essential hypertension) to 0.9089 (acute cerebrovascular disease). Unsurprisingly, chronic diseases are harder to predict than the acute ones (0.7475 vs 0.7964). 

We did not detect any positive correlation between disease prevalence and ROC-AUC score.
Moreover, the worst performance is observed for the most common phenotype (essential hypertension).

\begin{table}[ht]
\caption{ICU phenotypes used in the benchmark data set along with their prevalence and the per-phenotype classification performance of the best LSTM network}
\label{tab:phenotypes}
\centering
\begin{tabular}{llccc}
\toprule
Phenotype & Type & \multicolumn{2}{c}{Prevalence} & AUC-ROC \\
 &  & Train & Test \\
\midrule
Acute and unspecified renal failure                                      & acute   & 0.214 & 0.212 & 0.806 \\
Acute cerebrovascular disease                                            & acute   & 0.075 & 0.066 & 0.909 \\
Acute myocardial infarction                                              & acute   & 0.103 & 0.108 & 0.776 \\
Cardiac dysrhythmias                                                     & mixed   & 0.321 & 0.323 & 0.687 \\
Chronic kidney disease                                                   & chronic & 0.134 & 0.132 & 0.771 \\
Chronic obstructive pulmonary disease                                    & chronic & 0.131 & 0.126 & 0.695 \\
Complications of surgical/medical care                                   & acute   & 0.207 & 0.213 & 0.724 \\
Conduction disorders                                                     & mixed   & 0.072 & 0.071 & 0.737 \\
Congestive heart failure; nonhypertensive                                & mixed   & 0.268 & 0.268 & 0.763 \\
Coronary atherosclerosis and related                                     & chronic & 0.322 & 0.331 & 0.797 \\
Diabetes mellitus with complications                                     & mixed   & 0.095 & 0.094 & 0.872 \\
Diabetes mellitus without complication                                   & chronic & 0.193 & 0.192 & 0.797 \\
Disorders of lipid metabolism                                            & chronic & 0.291 & 0.289 & 0.728 \\
Essential hypertension                                                   & chronic & 0.419 & 0.423 & 0.683 \\
Fluid and electrolyte disorders                                          & acute   & 0.269 & 0.265 & 0.739 \\
Gastrointestinal hemorrhage                                              & acute   & 0.072 & 0.079 & 0.751 \\
Hypertension with complications                                          & chronic & 0.133 & 0.130 & 0.750 \\
Other liver diseases                                                     & mixed   & 0.089 & 0.089 & 0.778 \\
Other lower respiratory disease                                          & acute   & 0.051 & 0.057 & 0.694 \\
Other upper respiratory disease                                          & acute   & 0.040 & 0.043 & 0.785 \\
Pleurisy; pneumothorax; pulmonary collapse                               & acute   & 0.087 & 0.091 & 0.709 \\
Pneumonia                                                                & acute   & 0.139 & 0.135 & 0.809 \\
Respiratory failure; insufficiency; arrest                               & acute   & 0.181 & 0.177 & 0.907 \\
Septicemia (except in labor)                                             & acute   & 0.143 & 0.139 & 0.854 \\
Shock                                                                    & acute   & 0.078 & 0.082 & 0.892 \\
\midrule
All acute diseases (macro-averaged) & & & & 0.796\\
All mixed (macro-averaged) & & & & 0.768\\
All chronic diseases (macro-averaged) & & & & 0.746\\ 
\midrule
All diseases (macro-averaged) & & & & 0.776\\
\bottomrule
\end{tabular}
\end{table}



\subsection*{Multitask learning}
We demonstrated that the proposed multitask learning architecture allows us to extract certain useful information from the input sequence that single-task models could not leverage, which explains the better performance of multitask LSTM in some settings. 
We did not, however, find any significant benefit in using multitask learning for the phenotyping task.

We are interested in further investigating the practical challenges of multitask training.
In particular, for our four very different tasks, the model converges and then overfits at very different rates during training.
This is often addressed through the use of heuristics, including a multitask variant of early stopping, in which we identify the best epoch for each task based on individual task validation loss.
We proposed the use of per-task loss weighting, which reduced the problem but did not fully mitigate it.
One promising direction is to dynamically adapt these coefficients during training, similar to the adaptation of learning rates in optimizers.


\section*{Methods}
\label{sec:methods}

This section consists of three subsections. We describe the process of benchmark data and task generation along with evaluation metrics in the first subsection. The second subsection describes the linear and neural baseline models for the benchmark tasks. We describe the experimental setup and model selection in the third subsection. 

\subsection*{Benchmark tasks}

We first define some terminology: in MIMIC-III \textit{patients} are often referred to as \textit{subjects}.
Each patient has one or more \textit{hospital admissions}.
Within one admission, a patient may have one or more \textit{ICU stays}, which we also refer to as \textit{episodes}.
A clinical \textit{event} is an individual measurement, observation, or treatment.
In the context of our final task-specific data sets, we use the word \textit{sample} to refer to an individual record processed by a machine learning model.
As a rule, we have one sample for each prediction.
For tasks like phenotyping, a sample consists of an entire ICU stay.
For tasks requiring hourly predictions, e.g., LOS, a sample includes all events that occur before a specific time, and so a single ICU stay yields multiple samples.

\begin{figure}[tb!]
\includegraphics[width=0.99\textwidth]{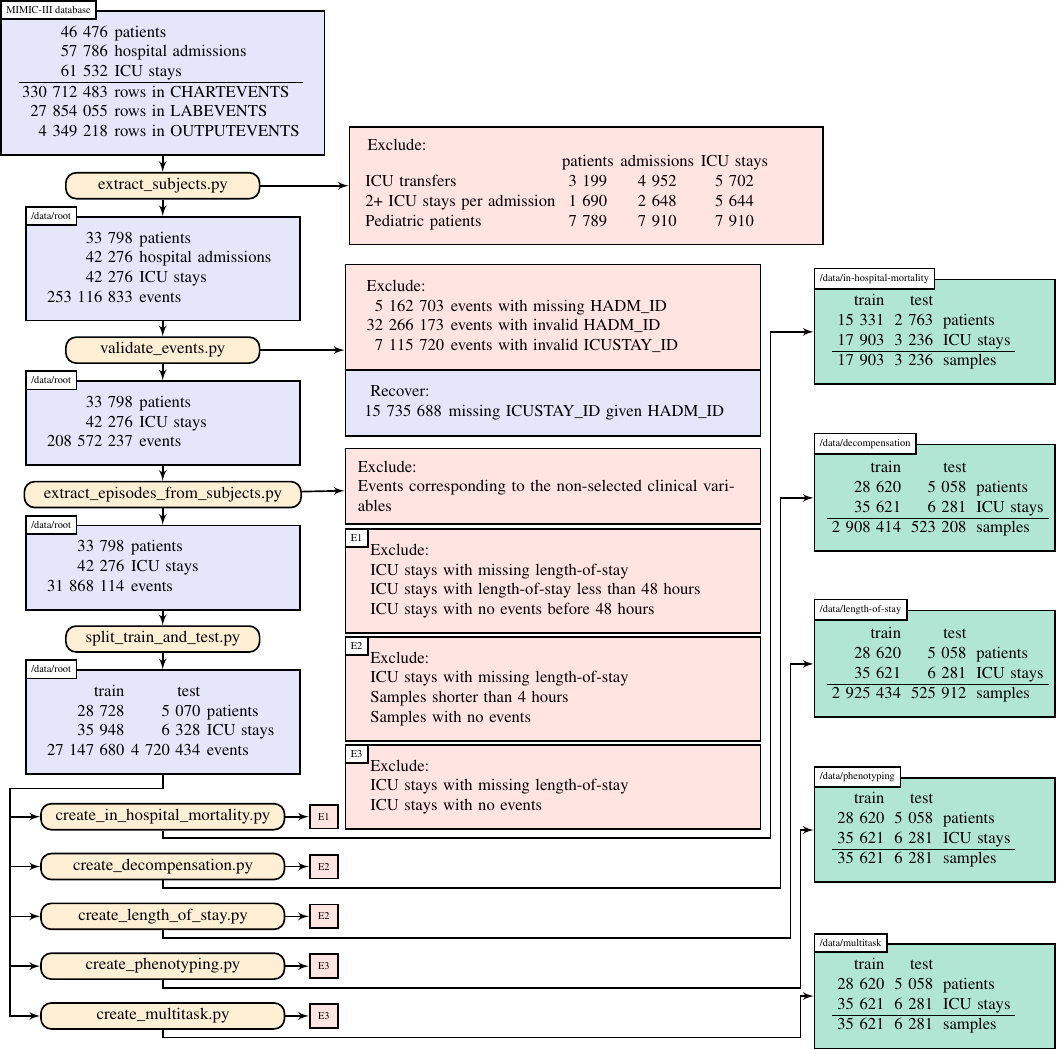}
\caption{Benchmark generation process.}
\label{fig:flowchart}
\end{figure}

Our benchmark preparation workflow, illustrated in Figure~\ref{fig:flowchart}, begins with the full MIMIC-III critical care database, which includes over 60,000 ICU stays across 40,000 critical care patients.
In the first step (\texttt{extract\_subjects.py}), we extract relevant data from the raw MIMIC-III tables and organize them by patient.
We also apply exclusion criteria to admissions and ICU stays.
First, we exclude any hospital admission with multiple ICU stays or transfers between different ICU units or wards.
This reduces the ambiguity of outcomes associated with hospital admissions rather than ICU stays.
Second, we exclude all ICU stays where the patient is younger than 18 due to the substantial differences between adult and pediatric physiology.
The resulting \textit{root} cohort has 33,798 unique patients with a total of 42,276 ICU stays and over 250 million clinical events.

In the next two steps, we process the clinical events.
The first step, \texttt{validate\_events.py}, filters out 45 million events that cannot be reliably matched to an ICU stay in our cohort.
First, it removes all events for which admission ID (HADM\_ID) is not present.
Then it excludes events that have admission ID that is not present in the \texttt{stays.csv} database which connects ICU stay properties (e.g. length of stay and mortality) to admissions IDs.
Next, it considers events with missing ICU stay IDs (ICUSTAY\_ID). 
For all such events the ICUSTAY\_ID is reliably recovered by looking at the HADM\_ID.
Finally, the script excludes all events for which the ICUSTAY\_ID is not listed in the \texttt{stays.csv} database. More detailed description of \texttt{validate\_events.py} script can be found in the code repository \cite{mimicbenchmarkrepo}.

The second step, \texttt{extract\_episodes\_from\_subjects.py}, compiles a time series of events for each episode, retaining only the variables from a predefined list and performing further cleaning.
We use 17 physiologic variables representing a subset from the Physionet/CinC Challenge 2012~\cite{silva2012physionet}. The selected variables are listed in the first column of Table \ref{tab:variables}. The data for each of these variables is compiled from multiple MIMIC-III variables. For example, there are eight variables in \texttt{chartevents} table that correspond to weight (denoted by item IDs 763, 224639, 226512, 3580, 3693, 3581, 226531 and 3582). These variables come in different units (kg, oz, lb). We convert all values to kilograms. We do similar preprocessing for all 17 variables. The mapping of MIMIC-III item IDs to our variables is available at \texttt{resources/itemid\_to\_variable\_map.csv} file of our code repository\cite{mimicbenchmarkrepo}, where the first column is the name of the variable in our benchmark.
The script \texttt{preprocessing.py} contains functions for converting these values to a unified scale.

The resulting data has over 31 million events from 42,276 ICU stays.

\begin{table}[thb!]
    \centering
    \begin{tabular}{llll}
    \toprule
    Variable & MIMIC-III table & Impute value & Modeled as \\
    \midrule
        Capillary refill rate & chartevents & 0.0 & categorical \\
        Diastolic blood pressure & chartevents & 59.0 & continuous \\
        Fraction inspired oxygen & chartevents & 0.21 & continuous \\
        Glascow coma scale eye opening & chartevents & 4 spontaneously & categorical \\
        Glascow coma scale motor response & chartevents & 6 obeys commands & categorical \\
        Glascow coma scale total & chartevents & 15 & categorical \\
        Glascow coma scale verbal response & chartevents & 5 oriented & categorical\\
        Glucose & chartevents, labevents & 128.0 & continuous \\
        Heart Rate & chartevents & 86 & continuous \\
        Height & chartevents & 170.0 & continuous \\
        Mean blood pressure & chartevents & 77.0 & continuous\\
        Oxygen saturation & chartevents, labevents & 98.0 & continuous\\
        Respiratory rate & chartevents & 19 & continuous\\
        Systolic blood pressure & chartevents & 118.0  & continuous\\
        Temperature & chartevents & 36.6 & continuous\\
        Weight & chartevents & 81.0 & continuous\\
        pH & chartevents, labevents & 7.4 & continuous\\
    \bottomrule
    \end{tabular}
    \caption{The 17 selected clinical variables. The second column shows the source table(s) of a variable from MIMIC-III database. The third column lists the ``normal'' values we used in our baselines during the imputation step, and the fourth column describes how our LSTM-based baselines treat the variables.}
    \label{tab:variables}
\end{table}

Finally (\texttt{split\_train\_and\_test.py}), we fix a test set of 15\% (5,070) of patients, including 6,328 ICU stays and 4.7 million events.
We encourage researchers to follow best practices by interacting with the test data as infrequently as possible.
Finally, we prepare the task-specific data sets.

Our benchmark prediction tasks include four in-hospital clinical prediction tasks: modeling risk of mortality shortly after admission~\cite{zimmerman2006acute}, real-time prediction of physiologic decompensation~\cite{williams2012national}, continuous forecasting of patient LOS~\cite{dahl2012high}, and phenotype classification~\cite{lipton2016learning}.
Each of these tasks is of interest to clinicians and hospitals and is directly related to one or more opportunities for transforming health care using big data~\cite{bates2014big}.
These clinical problems also encompass a range of machine learning tasks, including binary and multilabel classification, regression, and time series modeling, and so are of interest to data mining researchers.

\subsubsection*{In-hospital mortality}

Our first benchmark task involves prediction of in-hospital mortality from observations recorded early in an ICU admission.
Mortality is a primary outcome of interest in acute care:
ICU mortality rates are the highest among hospital units (10\% to 29\% depending on age and illness), and early detection of at-risk patients is key to improving outcomes.

Interest in modeling risk of mortality in hospitalized patients dates back over half a century: the Apgar score~\cite{apgar1952proposal} for assessing risk in newborns
was first published in 1952, the widely used Simplified Acute Physiology Score (SAPS)~\cite{le1984simplified} in 1984.
Intended to be computed by hand, these scores are designed to require as few inputs as possible and focus on individual abnormal observations rather than trends.
However, in the pursuit of increased accuracy, such scores have grown steadily more complex: the Acute Physiology and Chronic Health Evaluation (APACHE) IV score requires nearly twice as many clinical variables as APACHE II~\cite{zimmerman2006acute}.

Recent research has used machine learning techniques like state space models and time series mining to integrate
complex temporal patterns
instead of individual measurements~\cite{marlin2012unsupervised,luo2016predicting}.
Others leverage information from clinical notes, extracted using topic models~\cite{caballero2015dynamically,ghassemi2015multivariate}.
These approaches outperform traditional baselines but have not been compared on standardized benchmarks.

Risk of mortality is most often formulated as binary classification using observations recorded from a limited window of time following admission.
The target label indicates whether the patient died before hospital discharge.
Typical models include only the first 12-24 hours, but we use a wider 48-hour window to enable the detection of patterns that may indicate changes in patient acuity, similar to the PhysioNet/CinC Challenge 2012~\cite{silva2012physionet}.

The most commonly reported metric in mortality prediction research is area under the receiver operator characteristic curve (AUC-ROC).
We also report area under the precision-recall curve (AUC-PR) metric since it can be more informative when dealing with highly skewed datasets~\cite{Davis2006-roc-pr}.

To prepare our \textit{in-hospital-mortality} data set, we begin with the \textit{root} cohort and further exclude all ICU stays for which LOS is unknown or less than 48 hours or for which there are no observations in the first 48 hours.
This yields final training and test sets of 17,903 and 3,236 ICU stays, respectively.
We determined in-hospital mortality by comparing patient date of death with hospital admission and discharge times.
The resulting mortality rate is 13.23\% (2,797 of 21,139 ICU stays).

\subsubsection*{Physiologic Decompensation}

Our second benchmark task involves the detection of patients who are physiologically decompensating, or whose conditions are deteriorating rapidly.
Such patients are the focus of ``track-and-trigger'' initiatives~\cite{bates2014big}.
In such programs, patients with abnormal physiology trigger an alert, summoning a rapid response from a team of
specialists who assume care of the triggering patient.

These programs are typically implemented using early warning scores, which summarize patient state with a composite score and trigger alerts based on abnormally low values.
Examples include the Modified Early Warning Score (MEWS)~\cite{subbe2001validation}, the VitalPAC Early Warning Score (ViEWS)~\cite{prytherch2010views}, and the National Early Warning Score (NEWS)~\cite{williams2012national} being deployed throughout the United Kingdom.
Like risk scores, most early warning scores are designed to be computed manually and so are based on simple thresholds and a small number of common vital signs.

Detection of decompensation is closely related to problems like condition monitoring~\cite{aleks2009probabilistic}
and sepsis detection~\cite{henry2015targeted} that have received significant attention from the machine learning community.
In contrast, decompensation has seen relatively little research,
with one notable exception, where Gaussian process was used to impute missing values, enabling the continuous application of early warning scores even when vitals are not recorded~\cite{clifton2012gaussian}.

There are many ways to define decompensation, but most objective evaluations of early warning scores are based on accurate prediction of mortality within a fixed time window, e.g., 24 hours, after assessment~\cite{williams2012national}.
Following suit, we formulate our decompensation benchmark task as a binary classification problem, in which the target label indicates whether the patient dies within the next 24 hours.

To prepare the \textit{root} cohort for decompensation detection, we define a binary label that indicates whether the patient's date of death falls within the next 24 hours of the current time point.
We then assign these labels to each hour, starting at four hours after admission to the ICU (in order to avoid having too short samples) and ending when the patient dies or is discharged.
This yields 2,908,414 and 523,208 instances (individual time points with a label) in the training and test sets, respectively.
The decompensation rate is 2.06\% (70,696 out of 3.431,622 instances).

We use the same metrics for decompensation as for mortality, i.e., AUC-ROC and AUC-PR.
Because we care about per-instance (vs. per-patient) accuracy in this task, overall performance is computed as the micro-average over all predictions, regardless of patient.

\subsubsection*{Forecasting length of stay}

\begin{figure}[t]
\includegraphics[width=\textwidth]{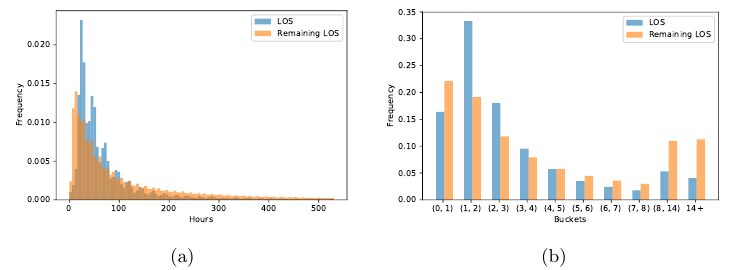}
\caption{Distribution of length of stay (LOS). (a) The distribution of LOS for full ICU stays and remaining LOS per hour. The rightmost 5\% of both distributions are not shown to keep the plot informative. (b) Histogram of bucketed patient and hourly remaining LOS (less than one day, one each for 1-7 days, between 7 and 14 days, and over 14 days).}
\label{fig:los_stats}
\end{figure}

Our third benchmark task involves forecasting hospital LOS, one of the most important drivers of overall hospital cost~\cite{AHRQ,dahl2012high}.
Hospitals use patient LOS as both a measure of a patient's acuity and for scheduling and resource management~\cite{AHRQ}.
Patients with extended LOS utilize more hospital resources and often have complex, persistent conditions that may not be immediately life threatening but are nonetheless difficult to treat.
Reducing health care spending requires early identification and treatment of such patients.

Most LOS research has focused on identifying factors that influence LOS~\cite{higgins2003early} rather than predicting it.
Both severity of illness scores~\cite{osler1998predicting} and early warning scores~\cite{paterson2006prediction} have been used to predict LOS but with mixed success.
There has been limited machine learning research concerned with LOS, most of it focused on specific conditions~\cite{pofahl1998use} and cohorts~\cite{grigsby1994simulated}.

LOS is naturally formulated as a regression task.
Traditional research focuses on accurate prediction of LOS early in admission, but in our benchmark we predict the \textit{remaining} LOS once per hour for every hour after admission, similar to decompensation.
Such a model can be used to help hospitals and care units make decisions about staffing and resources on a regular basis, e.g., at the beginning of each day or at shift changes.

We prepare the \textit{root} cohort for LOS forecasting in a manner similar to decompensation: for each time point, we assign a LOS target to each time point in sliding window fashion, beginning four hours after admission to the ICU and ending when the patient dies or is discharged.
We compute remaining LOS by subtracting total time elapsed from the existing LOS field in MIMIC-III.

After filtering, there remain 2,925,434 and 525,912 instances (individual time points) in the training and test sets, respectively.
Figure \ref{fig:los_stats} (a) shows the distributions of patient LOS and hourly remaining LOS in our final cohort.
Because there is no widely accepted evaluation metric for LOS predictions we use a standard regression metric -- mean absolute difference (MAD).

In practice, hospitals round to the nearest day when billing, and stays over 1-2 weeks are considered extreme outliers which, if predicted, would trigger special interventions~\cite{dahl2012high}.
Thus, we also design a custom metric that captures how LOS is measured and studied in practice.
First, we divide the range of values into ten buckets, one bucket for extremely short visits (less than one day), seven day-long buckets for each day of the first week, and two ``outlier'' buckets -- one for stays of over one week but less than two, and one for stays of over two weeks.
This converts length-of-stay prediction into an ordinal multiclass classification problem.
To evaluate prediction accuracy for this problem formulation, we use Cohen's linear weighted kappa~\cite{cohen1960coefficient,Cohen-uses-brennan1981coefficient}, which measures correlation between ordered items.
Figure \ref{fig:los_stats} (b) shows the distribution of bucketed LOS and hourly remaining LOS.

\subsubsection*{Acute care phenotype classification}

Our final benchmark task is phenotyping, i.e., classifying which acute care conditions are present in a given patient record. 
Phenotyping has applications in cohort construction for clinical studies, comorbidity detection and risk adjustment, quality improvement and surveillance, and diagnosis~\cite{oellrich2015digital}.
Traditional research phenotypes are identified via chart review based on criteria predefined by experts,
while surveillance phenotypes use simple definitions based primarily on billing, e.g., ICD-9, codes.
The adoption of EHRs has led to increased interest in machine learning approaches to phenotyping that treat it as classification~\cite{agarwal2016noisy,halpern2016anchors} or clustering~\cite{marlin2012unsupervised,ho2014mhp}

In this task we classify 25 conditions that are common in adult ICUs, including 12 critical (and sometimes life-threatening) conditions, such as respiratory failure and sepsis; eight chronic conditions that are common comorbidities and risk factors in critical care, such as diabetes and metabolic disorders; and five conditions considered ``mixed'' because they are recurring or chronic with periodic acute episodes.
To identify these conditions, we use the single-level definitions from the Health Cost and Utilization (HCUP) Clinical Classification Software (CCS)~\cite{CCS}.
These definitions group ICD-9 billing and diagnostic codes into mutually exclusive, largely homogeneous disease categories, reducing some of the noise, redundancy, and ambiguity in the original ICD-9 codes.
HCUP CCS code groups are used for reporting to state and national agencies, so they constitute sensible phenotype labels.

We determined phenotype labels based on the MIMIC-III ICD-9 diagnosis table. First, we mapped each code to its HCUP CCS category, retaining only the 25 categories from Table \ref{tab:phenotypes}.
We then matched diagnoses to ICU stays using the hospital admission identifier, since ICD-9 codes in MIMIC-III are associated with hospital visits, not ICU stays.
By excluding hospital admissions with multiple ICU stays, we reduced some of the ambiguity in these labels: there is only one ICU stay per hospital admission with which the diagnosis can be associated. 
Note that we perform ``retrospective'' phenotype classification, in which we observe a full ICU stay before predicting which diseases are present.
This is due in part to a limitation of MIMIC-III: the source of our disease labels, ICD-9 codes, do not have timestamps, so we do not know with certainty when the patient was diagnosed or first became symptomatic.
Rather than attempt to assign timestamps using a heuristic, we decided instead to embrace this limitation.
We apply no additional filtering to the phenotyping cohort, so there are 35,621 and 6,281 ICU stays in the training and test sets, respectively.
The full list of phenotypes is shown in Table \ref{tab:phenotypes}, along with prevalence within the benchmark data set.

Because diseases can co-occur (in fact, 99\% of patients in our benchmark data set have more than one diagnosis), we formulate phenotyping as a multi-label classification problem.
Similar to Lipton et al.~\cite{lipton2016learning}, we report macro- and micro-averaged AUC-ROC with the macro-averaged score being the main score.

\subsection*{Baselines}

In this subsection, we discuss two sets of models that we evaluate on each of our four benchmark tasks: linear (logistic) regression with hand-engineered features and LSTM-based neural networks.
Both have been shown to be effective for clinical prediction from physiologic time series.
For linear models, we briefly describe our feature engineering, which is also implemented in our benchmark code.
For LSTMs, we review the basic definition of the LSTM architecture, our data preprocessing, and the loss function for each task.
We then describe an atypical channel-wise variant of the LSTM that processes each variable separately, a deep supervision training strategy, and finally our heterogeneous multitask architecture.

\subsubsection*{Logistic regression}

For our logistic regression baselines, we use a more elaborate version of the hand-engineered features described in Lipton et al.~\cite{lipton2016learning}.
For each variable, we compute six different sample statistic features on seven different subsequences of a given time series.
The per-subsequence features include minimum, maximum, mean, standard deviation, skew and number of measurements.
The seven subsequences include the full time series, the first 10\% of time, first 25\% of time, first 50\% of time, last 50\% of time, last 25\% of time, last 10\% of time.
When a sub-sequence does not contain a measurement, the features corresponding to that sub-sequence are marked missing.
Notably, all categorical variables have numerical values with a meaningful scale.
Thus, we use no additional encoding of categorical variables.
In total, we obtain $17 \times 7 \times 6 = 714$ features per time series.
The missing values are replaced are with mean values computed on the training set.
Then the features are standardized by subtracting the mean and dividing by the standard deviation. 
We train a separate logistic regression classifier for each of mortality, decompensation, and the 25 phenotypes.
For LOS, we trained a softmax regression model to solve the 10-class bucketed LOS problem.

\subsubsection*{LSTM-based models}
We begin by briefly revisiting the fundamentals of long short-term memory recurrent neural networks (LSTM RNNs)~\cite{LSTM-hochreiter1997} and introducing notation for benchmark prediction tasks in order to describe our LSTM-based models.
The LSTM is a type of RNN designed to capture long term dependencies in sequential data.
It takes a sequence $\{x_t\}_{t \geq 1}^T$ of length $T$ as its input and outputs a $T$-long sequence of $\{h_t\}_{t \geq 1}^T$ hidden state vectors using the following equations:
\begin{align*}
i_t &= \sigma(x_t W^{(xi)} + h_{t-1} W^{(hi)} ),\\
f_t &= \sigma(x_t W^{(xf)} + h_{t-1} W^{(hf)} ),\\
c_t &= f_t \odot c_{t - 1}
       + i_t \odot \tanh(x_t W^{(xc)} + h_{t-1} W^{(hc)} + b^{(c)}),\\
o_t &= \sigma(x_t W^{(xo)} + h_{t-1} W^{(ho)} + b^{(o)}),\\
h_t &= o_t \odot \sigma_h(c_t),
\end{align*}
where $h_0 = 0$. The $\sigma$ (sigmoid) and $\tanh$ functions are applied element-wise.
We do not use peephole connections~\cite{peephole-gers2000recurrent}. The $W$ matrices and $b$ vectors are the trainable parameters of the LSTM.
Later we will use $h_t = LSTM(x_t, h_{t-1})$ as a shorthand for the above equations.
We apply dropout on non-recurrent connections between LSTM layers and before outputs.

For LSTM-based models we re-sample the time series into regularly spaced intervals.
If there are multiple measurements of the same variable in the same interval, we use the value of the last measurement.
We impute the missing values using the most recent measurement value if it exists and a pre-specified ``normal'' value otherwise (see the third column of Table \ref{tab:variables}).
In addition, we also provide a binary mask input for each variable indicating the timesteps that contain a true (vs. imputed) measurement~\cite{lipton2016missing}.
Categorical variables (even binary ones) are encoded using a one-hot vector.
Numeric inputs are standardized by subtracting the mean and dividing by the standard deviation.
The statistics are calculated per variable after imputation of missing values.

After the discretization and standardization steps we get 17 pairs of time series for each ICU stay: $(\{\mu^{(i)}_t\}_{t \geq 1}^T, \{c^{(i)}_t\}_{t \geq 1}^T)$, where $\mu^{(i)}_t$ is a binary variable indicating whether variable $i$ was observed at time step $t$ and $c^{(i)}_t$ is the value (observed or imputed) of variable $i$ at time step $t$.
By $\{x_t\}_{t \geq 1}^T$ we denote the concatenation of all $\{\mu^{(i)}_t\}_{t \geq 1}^T$ and $\{c^{(i)}_t\}_{t \geq 1}^T$ time series, where concatenation is done across the axis of variables.
In all our experiments $x_t$ becomes a vector of length 76.

We also have a set of targets for each stay: $\{d_t\}_{t \geq 1}^T$ where $d_t \in \{0,1\}$ is a list of $T$ binary labels for decompensation, one for each hour;
$m \in \{0, 1\}$ is single binary label indicating whether the patient died in-hospital;
$\{\ell_t\}_{t\geq1}^{T}$ where $\ell_t \in \mathbb{R}$ is a list of real valued numbers indicating \textit{remaining} length of stay (hours until discharge) at each time step; and $p_{1:K} \in \{0,1\}^{K}$ is a vector of $K$ binary phenotype labels.
When training our models to predict length of stay, we instead use a set of categorical labels $\{l_t\}_{t\geq1}^{T}$ where $l_t \in \{1, \dots, 10\}$ indicates in which of the ten length-of-stay buckets $\ell_t$ belongs.
When used in the context of equations (e.g., as the output of a softmax or in a loss function), we will interpret $l_t$ as a one-of-ten hot binary vector, indexing the $i$th entry as $l_{ti}$.

Note that because of task-specific filters are applied in the creation of benchmark tasks, we may have situations where for a given stay $m$ is missing and/or $d_t$, $\ell_t$ are missing for some time steps. Without abusing the notation in our equations we will assume that all targets are present. In the code missing targets are discarded.

We describe the notations of the instances for each benchmark task.
Each instance of in-hospital mortality prediction task is a pair $(\{x_t\}_{t \geq 1}^{48}, m)$, where $x$ is the matrix of clinical observations of first 48 hours of the ICU stay and $m$ is the label.
An instance of decompensation and length of stay prediction tasks is a pair $(\{x_t\}_{t \geq 1}^{\tau}, y)$, where $x$ is the matrix of clinical observations of first $\tau$ hours of the stay and $y$ is the target variable (either $d_{\tau}$, ${\ell}_{\tau}$ or $l_{\tau})$.
Each instance of phenotype classification task is a pair $(\{x_t\}_{t \geq 1}^{T}, p_{1:K})$, where $x$ is the matrix of observation of the whole ICU stay and $p_{1:K}$ are the phenotype labels.

Our first LSTM-based baseline takes an instance $(\{x_t\}_{t \geq 1}^T, y)$ of a prediction task and uses a single LSTM layer to process the input: $h_t = LSTM(x_t, h_{t-1})$. 
To predict the target we add the output layer:
\begin{align*}
\widehat{d_T} &= \sigma\left(w^{(d)}h_T+b^{(d)}\right), \\
\widehat{m} &= \sigma\left(w^{(m)}h_T+b^{(m)}\right), \\
\widehat{\ell_T} &= \mathrm{relu}\left(w^{(\ell)}h_T+b^{(\ell)}\right), \\
\widehat{l_T} &= \mathrm{softmax}\left(W^{(l)}h_T+b^{(l)}\right), \\
\widehat{p_i} &= \sigma\left(W^{(p)}_{i,\cdot}h_T+b^{(p)}_i\right),
\end{align*}
where $y$ is $d_T$, $m$, $\ell_T$, $l_T$ or $p_{1:K}$ respectively. 
The loss functions we use to train these models are (in the same order as above):
\begin{align*}
\mathcal{L}_d &= CE(d_T, \widehat{d_T}), \\
\mathcal{L}_m &= CE(m, \widehat{m}), \\
\mathcal{L}_{\ell} &= (\widehat{\ell_T} - {\ell}_T)^2, \\
\mathcal{L}_l &= MCE(l_T, \widehat{l_T}), \\
\mathcal{L}_p &= \frac{1}{K}\sum\limits_{k=1}^K{CE(p_k, \widehat{p_k}}),
\end{align*}
where $CE(y, \widehat{y})$ is the binary cross entropy and $MCE(y, \widehat{y})$ is multiclass cross entropy defined over the $C$ classes:
\begin{align*}
\mathrm{CE}(y, \widehat{y}) &= -\left(y \cdot \log(\widehat{y}) + (1-y)\cdot\log(1-\widehat{y}) \right), \\
\mathrm{MCE}(y, \widehat{y}) &= - \sum_{k=1}^{C} y_{k} \log(\widehat{y}_{k}).
\end{align*}
We call this model ''Standard LSTM``.

\subsubsection*{Channel-wise LSTM}

In addition to the standard LSTM baseline, we also propose a modified LSTM baseline which we call channel-wise LSTM.
While the standard LSTM network work directly on the concatenation $\{x_t\}_{t \geq 1}^T$ of the time series, the channel-wise LSTM pre-processes the data $(\{\mu^{(i)}_t\}_{t \geq 1}^T, \{c^{(i)}_t\}_{t \geq 1}^T)$ of different variables independently using a bidirectional LSTM layer.
We use different LSTM layers for different variables.
Then the outputs of these LSTM layers are concatenated and are fed to another LSTM layer.
\begin{align*}
p^{(i)}_t &= LSTM([\mu^{(i)}_t; c^{(i)}_t], p^{(i)}_{t-1}) \\
q^{(i)}_t &= LSTM([\overleftarrow{\mu^{(i)}}_t; \overleftarrow{c^{(i)}}_t], q^{(i)}_{t-1}) \\
u_t &= [p^{(1)}_t; \overleftarrow{q^{(1)}}_t; \ldots p^{(17)}_t; \overleftarrow{q^{(17)}}_t] \\
h_t &= LSTM(u_t, h_{t-1})
\end{align*}

$\overleftarrow{x}_t$ denotes the $t$-th element of the reverse of the sequence $\{x_t\}_{t \geq 1}^T$.
The output layers and loss functions for each task are the same as those in the standard LSTM baseline.

The intuition behind having channel-wise module is two-fold.
First, it helps to pre-process the data of a single variable before mixing it with the data of other variables.
This way the model can learn to store some useful information related to only that particular variable.
For example, it can learn to store the maximum heart rate or the average blood pressure in earlier time steps.
This kind of information is hard to learn in the case of standard LSTMs, as the input to hidden weight matrices need to have sparse rows.
Second, this channel-wise module facilitates incorporation of missing data information by explicitly showing which mask variables relate to which variables.
This information can be tricky to learn in standard LSTM models.
While in most of our experiments channel-wise LSTMs outperformed standard LSTMs, we did not perform an extensive ablation study to determine the contribution of various components of channel-wise LSTMs (grouping of related variables, bidirectionality, the additional LSTM layer) in the performance gain. 

Note that this channel-wise module can be used as a replacement of the input layer in any neural architecture which takes the concatenation of time series of different variables as its input.

\subsubsection*{Deep supervision}

So far we defined models that do the prediction in the last
step.
This way the supervision comes from the last time step, implying that the model needs to pass information across many time steps.
We propose two methods where we supervise the model at each time step.
We use the term deep supervision to refer them.

For in-hospital mortality and phenotype prediction tasks we use target replication~\cite{lipton2016learning} to do deep supervision.
In this approach we replicate the target in all time steps and by changing the loss function we require the model to predict the replicated target variable too. 
The loss functions of these deeply supervised models become:
\begin{align*}
\mathcal{L}^*_m &= (1-\alpha) * \mathrm{CE}(m, \widehat{m_T}) + \alpha * \frac{1}{T}\sum\limits_{t=1}^T{\mathrm{CE}(m, \widehat{m_t})}, \\
\mathcal{L}^*_p &= \frac{1}{K} \sum\limits_{i=1}^{K}{\left((1-\alpha) *  \mathrm{CE}(p_k, \widehat{p}_{T,k}) + \alpha * \frac{1}{T}\sum\limits_{t=1}^T{\mathrm{CE}(p_k, \widehat{p}_{tk})} \right)},
\end{align*}
where $\alpha \in [0, 1] $ is a hyperparameter that represents the strength of target replication part in loss functions, $\widehat{d_t}$ is decompensation prediction at time step $t$, and $\widehat{p_{tk}}$ is the prediction of $k$-th phenotype a time step $t$.

For decompensation and length of stay prediction tasks we cannot use target replication, because the target of the last time step might be wrong for the other time steps.
Since in these tasks we create multiple prediction instances from a single ICU stay, we can group these samples and predict them in a single pass. 
This way we will have targets for each time step and the model will be supervised at each time step.
The loss functions of these deeply supervised models are:
\begin{align*}
\mathcal{L}^*_d &= \frac{1}{T} \sum\limits_{t=1}^{T} \mathrm{CE}(d_t, \widehat{d}_t), \\
\mathcal{L}^*_{\ell} &= \frac{1}{T} \sum\limits_{t=1}^{T} (\widehat{\ell_t} - \ell_t)^2, \\
\mathcal{L}^*_l &= \frac{1}{T} \sum\limits_{t=1}^{T} \mathrm{MCE}(l_{t}, \widehat{l_t}). \\
\end{align*}

Note that whenever we group the instances of a single ICU stay, we use simple left-to-right LSTMs instead of bidirectional LSTMs, so that the data from future time steps is not used.

\subsubsection*{Multitask learning LSTM}

\begin{figure}[t]
{\includegraphics[width=\textwidth]{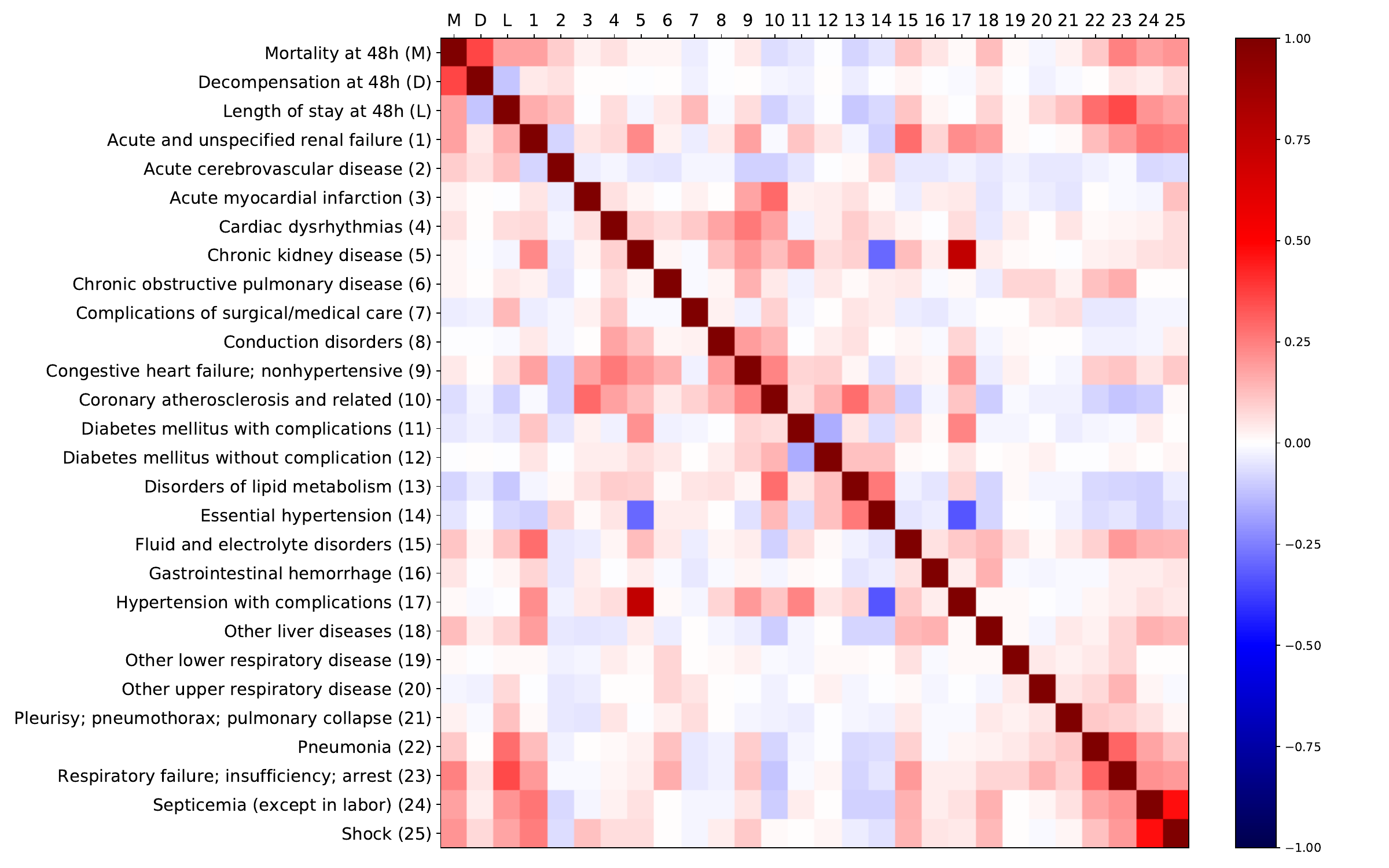}}
\caption{Correlations between task labels.}
\label{fig:correlations}
\end{figure}

So far we predicted targets for each task independently.
There is a natural question about the effectiveness of multitasking.
Correlations between the targets of different tasks are presented in the Figure \ref{fig:correlations}.
For each task we propose another baseline, where we try to use the other three tasks as auxiliary tasks to enhance the performance.
This multitasking can be done with either standard LSTM or channel-wise LSTM.

In multitask setting we group the instances coming from a single ICU stay and predict all targets associated with a single ICU jointly. 
This means we use deep supervision of decompensation and length of stay prediction tasks. We are free to choose whether we want to use deep supervision for in-hospital mortality and phenotype prediction tasks. 

\begin{figure}[t]
\begin{center}
\includegraphics[width=0.8\textwidth]{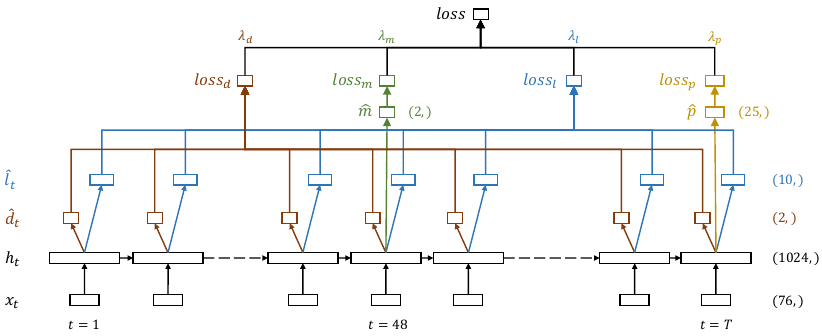}
\end{center}
\caption{LSTM-based network architecture for multitask learning.}
\label{multitask-network}
\end{figure}

For in-hospital mortality, we consider only the first 48 timesteps $\{x_t\}_{t \geq 1}^{t_m}$, and predict $\widehat{m}$ at $t_m=48$ by adding a single dense layer with sigmoid activation which takes $h_{t_{m}}$ as its input.
For decompensation, we take the full sequence $\{x_t\}_{t \geq 1}^T$ and generate a sequence of mortality predictions $\{\widehat{d}\}_{t \geq 1}^T$ by adding a dense layer at every step.
For phenotyping, we consider the full sequence but predict phenotypes $\widehat{p}$ only at the last timestep $T$ by adding 25 parallel dense layers with sigmoid activations.
Similar to decompensation, we predict LOS by adding a single dense layer at each timestep. We experiment with two settings. In one case the dense layer outputs a single real number $\widehat{\ell}_t$, in the other case it uses softmax activation to output a distribution over the ten LOS buckets $\widehat{l}_t$.
The full multitask LSTM architecture is illustrated in Figure \ref{multitask-network}.

The loss functions for each task are the same as those in deep supervised setting. The overall loss is a weighted sum of task-specific losses:
$$
\mathcal{L}_{mt} = \lambda_d \cdot \mathcal{L}^*_d + \lambda_l \cdot \mathcal{L}^*_l + \lambda_m \cdot \mathcal{L}^*_m + \lambda_p \cdot \mathcal{L}^*_p,
$$
where the weights are non-negative numbers.
For raw length of stay prediction we replace $\mathcal{L}^*_l$ with $\mathcal{L}^*_{\ell}$ in the multitasking loss function.

\subsection*{Experiments, Model selection and Evaluation}

For all algorithms we use the data of the same 15\% patients of the predefined training set as validation data and train the models on the remaining 85\%.
We use grid search to tune all hyperparameters based on validation set performance. 
The best model for each baseline is chosen according to the performance on the validation set.
The final scores are reported on the test set, which we used sparingly during model development in order to avoid unintentional test set leakage. 

The only hyperparameters of logistic regression models are the coefficients of L1 and L2 regularizers.
In our experiments L1 and L2 penalties are mutually exclusive and are not applied to bias terms.
We used grid search to find the best penalty type and its regularization coefficient. For L2 penalty the grid of inverse of regularization coefficient includes $10^{-5}, 10^{-4}, \ldots, 1$, while for L1 penalty it includes $10^{-4}, 10^{-3}, \ldots, 1$.
For in-hospital mortality and decompensation prediction, the best performing logistic regression used L2 regularization with $C=0.001$.
For phenotype prediction, the best performing logistic regression used L1 regularization with $C=0.1$.
For LOS prediction, the best performing logistic regression used L2 regularization with $C=10^{-5}$.

When discretizing the data for LSTM-based models, we set the length of regularly spaced intervals to 1 hour.
This gives a reasonable balance between amount of missing data and number of measurements of the same variable that fall into the same interval.
This choice also agrees with the rate of sampling prediction instances for decompensation and LOS prediction tasks.
We also tried to use intervals of length 0.8 hours, but there was no improvement in the results.
For LSTM-based models, hyperparameters include the number of memory cells in LSTM layers, the dropout rate, and whether to use one or two LSTM layers.
Channel-wise LSTM models have one more hyperparameter - the number of units in channel-wise LSTMs (all the 17 LSTMs having the same number of units).
Whenever the target replication is enabled, we set $\alpha = 0.5$ in the corresponding loss function.

The best values of hyperparameters of LSTM-based models vary across the tasks. They are listed in \texttt{pretrained\_models.md} file in our code repository\cite{mimicbenchmarkrepo}. Generally, we noticed that dropout helps a lot to reduce overfitting. In fact, all LSTM-based baselines for in-hospital mortality prediction task (where the problem of overfitting is the most severe) use 30\% dropout.

For multitask models we have 4 more hyperparameters: $\lambda_d$, $\lambda_m$, $\lambda_l$ and $\lambda_p$ weights in the loss function. 
We didn't do full grid search for tuning these hyperparameters.
Instead we tried 5 different values of $(\lambda_d, \lambda_m, \lambda_l, \lambda_p)$ : $(1, 1, 1, 1)$; $(4, 2.5, 0.3, 1)$; $(1, 0.4, 3, 1)$; $(1, 0.2, 1.5, 1)$ and $(0.1, 0.1, 0.5, 1)$.
The first has the same weighs for each task, while the second tries to make the four summands of the loss function approximately equal.
The three remaining combinations were selected by looking at the speeds of learning of each task.

Overfitting was a serious problem in multitasking setup, with mortality and decompensation prediction validation performance degrading faster than the others.
All of the best multitask baselines use either $(1, 0.2, 1.5, 1)$ or $(0.1, 0.1, 0.5, 1)$ for $(\lambda_d, \lambda_m, \lambda_l, \lambda_p)$.
The first configuration performed the best for in-hospital mortality, decompensation and length of stay prediction tasks, whereas the second configuration was better for phenotype prediction task.
The fact that $\lambda_d$, $\lambda_m$ and $\lambda_l$ of the best multitask baselines for phenotype prediction task are relatively small supports the hypothesis that additional multitasking in phenotype prediction task hurts the performance.

All LSTM-based models are trained using ADAM~\cite{kingma2014adam} with a $10^{-3}$ learning rate and $\beta_1=0.9$. The batch size is set to either 8, 16 or 32 depending on the memory available at the computational unit.
We did not do extensive experiments to check whether tuning the hyperparameters of the optimizer and the batch size improves performance.

Finally, we evaluate the best baselines on their corresponding test sets.
Since the test score is an estimate of model performance on unseen examples, we use bootstrapping to estimate confidence intervals of the score. 
Bootstrapping have been used to estimate the standard deviations of the evaluation measures~\cite{choi2016retain}, to compute statistically significant differences between different models~\cite{BioCreativeII} and to report 95\% bootstrap confidence intervals for the models~\cite{CheXNet,GoogleBrainEHR2018}.
Providing confidence intervals also helps us to fight against a known problem of all public benchmark datasets -- overfitting on the test set.
To estimate a 95\% confidence interval we resample the test set $K$ times; calculate the score on the resampled sets; and use 2.5 and 97.5 percentiles of these scores as our confidence interval estimate.
For in-hospital mortality and phenotype prediction $K$ is 10000, while for decompensation and length-of-stay prediction $K$ is 1000, since the test sets of these tasks are much bigger.

\subsection*{Code availability}
The code to reproduce the results is available on Zenodo repository \cite{mimicbenchmarkrepo}.

\subsection*{Data availability}
MIMIC-III database analyzed in this study is available on PhysioNet repository \cite{mimicdata}. The code to generate the datasets used in this study is available on Zenodo repository \cite{mimicbenchmarkrepo}.

\section*{Acknowledgements}
The authors would like to thank Zachary Lipton for helpful comments. H.H. and H.K. were partially funded by an ISTC research grant. G.V.S. and A.G. acknowledge support by the Defense Advanced Research Projects Agency (DARPA) under grant FA8750-17-C-0106.

\section*{Author contributions}
D.K. had the initial idea for the analysis. D.K., G.V.S. and A.G. designed the study.  H.H. and D.K. designed the data processing pipeline. H.H. and H.K. designed machine learning experiments. H.H. performed the experiments. H.H. H.K. and D.K. analyzed the results and drafted the paper. H.K., G.V.S. and A.G. revised the paper draft.


\section*{Competing interests} 
The authors declare no competing interests.


\begin{thebibliography}{10}
\urlstyle{rm}
\expandafter\ifx\csname url\endcsname\relax
  \def\url#1{\texttt{#1}}\fi
\expandafter\ifx\csname urlprefix\endcsname\relax\def\urlprefix{URL }\fi
\expandafter\ifx\csname doiprefix\endcsname\relax\def\doiprefix{DOI: }\fi
\providecommand{\bibinfo}[2]{#2}
\providecommand{\eprint}[2][]{\url{#2}}

\bibitem{healthcare2014introduction}
\newblock \bibinfo{journal}{\bibinfo{title}{\textit{Introduction to the HCUP National Inpatient Sample (NIS) 2012}}}.
\newblock (Agency for Healthcare Research and Quality, \bibinfo{year}{2014}).

\bibitem{henryadoption}
\bibinfo{author}{Henry, J.}, \bibinfo{author}{Pylypchuk, Y.},
  \bibinfo{author}{Talisha~Searcy, M.} \& \bibinfo{author}{Patel, V.}
\newblock \bibinfo{journal}{\bibinfo{title}{Adoption of electronic health
  record systems among US non-federal acute care hospitals: 2008-2015}}.
\newblock {\emph{\JournalTitle{ONC Data Brief}}} \textbf{\bibinfo{volume}{35}}
  (Office of the National Coordinator for Health Information Technology, Washington DC, USA, \bibinfo{year}{2015}).

\bibitem{bates2014big}
\bibinfo{author}{Bates, D.~W.}, \bibinfo{author}{Saria, S.},
  \bibinfo{author}{Ohno-Machado, L.}, \bibinfo{author}{Shah, A.} \&
  \bibinfo{author}{Escobar, G.}
\newblock \bibinfo{journal}{\bibinfo{title}{Big data in health care: using
  analytics to identify and manage high-risk and high-cost patients}}.
\newblock {\emph{\JournalTitle{Health Affairs}}} \textbf{\bibinfo{volume}{33}},
  \bibinfo{pages}{1123--1131} (\bibinfo{year}{2014}).

\bibitem{zimmerman2006acute}
\bibinfo{author}{Zimmerman, J.~E.}, \bibinfo{author}{Kramer, A.~A.},
  \bibinfo{author}{McNair, D.~S.} \& \bibinfo{author}{Malila, F.~M.}
\newblock \bibinfo{journal}{\bibinfo{title}{Acute physiology and chronic health
  evaluation (apache) iv: hospital mortality assessment for today’s
  critically ill patients}}.
\newblock {\emph{\JournalTitle{Crit. Care Med.}}}
  \textbf{\bibinfo{volume}{34}}, \bibinfo{pages}{1297--1310}
  (\bibinfo{year}{2006}).

\bibitem{williams2012national}
\bibinfo{author}{Williams, B.} \emph{et~al.}
\newblock \bibinfo{journal}{\bibinfo{title}{\textit{National Early Warning Score
  (NEWS): Standardising the assessment of acute-illness severity in the NHS}}}.
\newblock (London: The Royal College of Physicians, \bibinfo{year}{2012}).

\bibitem{dahl2012high}
\bibinfo{author}{Dahl, D.} \emph{et~al.}
\newblock \bibinfo{journal}{\bibinfo{title}{The high cost of low-acuity icu
  outliers}}.
\newblock {\emph{\JournalTitle{Journal of Healthcare Management}}}
  \textbf{\bibinfo{volume}{57}}, \bibinfo{pages}{421--433}
  (\bibinfo{year}{2012}).

\bibitem{saria2015subtyping}
\bibinfo{author}{Saria, S.} \& \bibinfo{author}{Goldenberg, A.}
\newblock \bibinfo{journal}{\bibinfo{title}{Subtyping: What it is and its role
  in precision medicine}}.
\newblock {\emph{\JournalTitle{IEEE Intelligent Systems}}}
  \textbf{\bibinfo{volume}{30}}, \bibinfo{pages}{70--75}
  (\bibinfo{year}{2015}).

\bibitem{iserson2007triage}
\bibinfo{author}{Iserson, K.~V.} \& \bibinfo{author}{Moskop, J.~C.}
\newblock \bibinfo{journal}{\bibinfo{title}{Triage in medicine, part i:
  concept, history, and types}}.
\newblock {\emph{\JournalTitle{Ann. Emerg. Med.}}}
  \textbf{\bibinfo{volume}{49}}, \bibinfo{pages}{275--281}
  (\bibinfo{year}{2007}).

\bibitem{apgar1952proposal}
\bibinfo{author}{Apgar, V.}
\newblock \bibinfo{journal}{\bibinfo{title}{A proposal for a new method of
  evaluation of the newborn}}.
\newblock {\emph{\JournalTitle{Curr. Res. Anesth. Analg.}}}
  \textbf{\bibinfo{volume}{32}}, \bibinfo{pages}{260--267}
  (\bibinfo{year}{1952}).

\bibitem{ferrucci2013watson}
\bibinfo{author}{Ferrucci, D.}, \bibinfo{author}{Levas, A.},
  \bibinfo{author}{Bagchi, S.}, \bibinfo{author}{Gondek, D.} \&
  \bibinfo{author}{Mueller, E.~T.}
\newblock \bibinfo{journal}{\bibinfo{title}{Watson: beyond jeopardy!}}
\newblock {\emph{\JournalTitle{Artificial Intelligence}}}
  \textbf{\bibinfo{volume}{199}}, \bibinfo{pages}{93--105}
  (\bibinfo{year}{2013}).

\bibitem{silver2016mastering}
\bibinfo{author}{Silver, D.} \emph{et~al.}
\newblock \bibinfo{journal}{\bibinfo{title}{Mastering the game of go with deep
  neural networks and tree search}}.
\newblock {\emph{\JournalTitle{Nature}}} \textbf{\bibinfo{volume}{529}},
  \bibinfo{pages}{484--489} (\bibinfo{year}{2016}).


\bibitem{caballero2015dynamically}
\bibinfo{author}{Caballero~Barajas, K.~L.} \& \bibinfo{author}{Akella, R.}
\newblock \bibinfo{title}{Dynamically modeling patient's health state from
  electronic medical records: A time series approach}.
\newblock In \emph{\bibinfo{booktitle}{Proceedings of the 21th ACM SIGKDD
  International Conference on Knowledge Discovery and Data Mining}},
  \bibinfo{pages}{69--78} (\bibinfo{publisher}{ACM}, \bibinfo{address}{Sydney,
  Australia}, \bibinfo{year}{2015}).

\bibitem{ghassemi2015multivariate}
\bibinfo{author}{Ghassemi, M.} \emph{et~al.}
\newblock \bibinfo{title}{A multivariate timeseries modeling approach to
  severity of illness assessment and forecasting in icu with sparse,
  heterogeneous clinical data}.
\newblock In \emph{\bibinfo{booktitle}{Proceedings of the Twenty-Ninth AAAI
  Conference on Artificial Intelligence}}, \bibinfo{pages}{446--453}
  (\bibinfo{publisher}{AAAI Press}, \bibinfo{address}{Austin, Texas},
  \bibinfo{year}{2015}).

\bibitem{luo2016predicting}
\bibinfo{author}{Luo, Y.}, \bibinfo{author}{Xin, Y.}, \bibinfo{author}{Joshi,
  R.}, \bibinfo{author}{Celi, L.} \& \bibinfo{author}{Szolovits, P.}
\newblock \bibinfo{title}{Predicting icu mortality risk by grouping temporal
  trends from a multivariate panel of physiologic measurements}.
\newblock In \emph{\bibinfo{booktitle}{Proceedings of the Thirtieth AAAI
  Conference on Artificial Intelligence}}, \bibinfo{pages}{42--50}
  (\bibinfo{publisher}{AAAI Press}, \bibinfo{address}{Phoenix, Arizona},
  \bibinfo{year}{2016}).

\bibitem{lee2017customization}
\bibinfo{author}{Lee, J.} \& \bibinfo{author}{Maslove, D.~M.}
\newblock \bibinfo{journal}{\bibinfo{title}{Customization of a severity of
  illness score using local electronic medical record data}}.
\newblock {\emph{\JournalTitle{Journal of intensive care medicine}}}
  \textbf{\bibinfo{volume}{32}}, \bibinfo{pages}{38--47}
  (\bibinfo{year}{2017}).

\bibitem{johnson2017mlhc}
\bibinfo{author}{Johnson, A.}, \bibinfo{author}{Pollard, T.} \&
  \bibinfo{author}{Mark, R.}
\newblock \bibinfo{title}{Reproducibility in critical care: a mortality
  prediction case study}.
\newblock In \emph{\bibinfo{booktitle}{Proceedings of the 2nd Machine Learning
  for Healthcare Conference}}, vol.~\bibinfo{volume}{68},
  \bibinfo{pages}{361--376} (\bibinfo{publisher}{PMLR, Boston, Massachusetts},
  \bibinfo{year}{2017}).

\bibitem{quinn2009factorial}
\bibinfo{author}{Quinn, J.~A.}, \bibinfo{author}{Williams, C.~K.} \&
  \bibinfo{author}{McIntosh, N.}
\newblock \bibinfo{journal}{\bibinfo{title}{Factorial switching linear
  dynamical systems applied to physiological condition monitoring}}.
\newblock {\emph{\JournalTitle{IEEE Trans. Pattern Anal. Mach. Intell.}}} \textbf{\bibinfo{volume}{31}},
  \bibinfo{pages}{1537--1551} (\bibinfo{year}{2009}).

\bibitem{Laxmisan2007801}
\bibinfo{author}{Laxmisan, A.} \emph{et~al.}
\newblock \bibinfo{journal}{\bibinfo{title}{The multitasking clinician:  Decision-making and cognitive demand during and after team handoffs in  emergency care}}.
\newblock {\emph{\JournalTitle{Int. J. Med. Inform.}}}
  \textbf{\bibinfo{volume}{76}}, \bibinfo{pages}{801 -- 811}
  (\bibinfo{year}{2007}).

\bibitem{horn1991relationship}
\bibinfo{author}{Horn, S.~D.} \emph{et~al.}
\newblock \bibinfo{journal}{\bibinfo{title}{The relationship between severity
  of illness and hospital length of stay and mortality.}}
\newblock {\emph{\JournalTitle{Med. Care}}} \textbf{\bibinfo{volume}{29}},
  \bibinfo{pages}{305--317} (\bibinfo{year}{1991}).

\bibitem{johnson2016mimic}
\bibinfo{author}{Johnson, A. E.~W.} \emph{et~al.}
\newblock \bibinfo{journal}{\bibinfo{title}{MIMIC-III, a freely accessible
  critical care database}}.
\newblock {\emph{\JournalTitle{Scientific Data}}} \textbf{\bibinfo{volume}{3}},
  \bibinfo{pages}{160035} (\bibinfo{year}{2016}).

\bibitem{mimicdata}
\bibinfo{author}{Laboratory For Computational~Physiology, M. I.~T.}
\newblock \bibinfo{journal}{\bibinfo{title}{The MIMIC-III clinical database}}.
\newblock {\emph{\JournalTitle{PhysioNet}}}
  \url{https://doi.org/10.13026/C2XW26} (\bibinfo{year}{2015}).

\bibitem{caruana1996using}
\bibinfo{author}{Caruana, R.}, \bibinfo{author}{Baluja, S.} \&
  \bibinfo{author}{Mitchell, T.}
\newblock \bibinfo{title}{Using the future to "sort out" the present: Rankprop
  and multitask learning for medical risk evaluation}.
\newblock In \emph{\bibinfo{booktitle}{Advances in Neural Information
  Processing Systems 8}}, \bibinfo{pages}{959--965} (\bibinfo{publisher}{MIT
  Press}, \bibinfo{address}{Denver, Colorado}, \bibinfo{year}{1996}).

\bibitem{clermont2001predicting}
\bibinfo{author}{Clermont, G.}, \bibinfo{author}{Angus, D.~C.},
  \bibinfo{author}{DiRusso, S.~M.}, \bibinfo{author}{Griffin, M.} \&
  \bibinfo{author}{Linde-Zwirble, W.~T.}
\newblock \bibinfo{journal}{\bibinfo{title}{Predicting hospital mortality for
  patients in the intensive care unit: a comparison of artificial neural
  networks with logistic fmultion models}}.
\newblock {\emph{\JournalTitle{Crit. Care Med.}}}
  \textbf{\bibinfo{volume}{29}}, \bibinfo{pages}{291--296}
  (\bibinfo{year}{2001}).

\bibitem{celi2012database}
\bibinfo{author}{Celi, L.~A.} \emph{et~al.}
\newblock \bibinfo{journal}{\bibinfo{title}{A database-driven decision support
  system: customized mortality prediction}}.
\newblock {\emph{\JournalTitle{J. Pers. Med.}}}
  \textbf{\bibinfo{volume}{2}}, \bibinfo{pages}{138--148}
  (\bibinfo{year}{2012}).

\bibitem{GoogleBrainEHR2018}
\bibinfo{author}{Rajkomar, A.} \emph{et~al.}
\newblock \bibinfo{journal}{\bibinfo{title}{Scalable and accurate deep learning
  with electronic health records}}.
\newblock {\emph{\JournalTitle{NPJ Digital Med.}}}
  \textbf{\bibinfo{volume}{1}}, \bibinfo{pages}{18} (\bibinfo{year}{2018}).

\bibitem{grigsby1994simulated}
\bibinfo{author}{Grigsby, J.}, \bibinfo{author}{Kooken, R.} \&
  \bibinfo{author}{Hershberger, J.}
\newblock \bibinfo{journal}{\bibinfo{title}{Simulated neural networks to
  predict outcomes, costs, and length of stay among orthopedic rehabilitation
  patients.}}
\newblock {\emph{\JournalTitle{Arch. Phys. Med. Rehabil.}}} \textbf{\bibinfo{volume}{75}}, \bibinfo{pages}{1077--1081}
  (\bibinfo{year}{1994}).

\bibitem{mobley1995artificial}
\bibinfo{author}{Mobley, B.~A.}, \bibinfo{author}{Leasure, R.} \&
  \bibinfo{author}{Davidson, L.}
\newblock \bibinfo{journal}{\bibinfo{title}{Artificial neural network
  predictions of lengths of stay on a post-coronary care unit}}.
\newblock {\emph{\JournalTitle{Heart \& Lung: The Journal of Acute and Critical
  Care}}} \textbf{\bibinfo{volume}{24}}, \bibinfo{pages}{251--256}
  (\bibinfo{year}{1995}).

\bibitem{buchman1994comparison}
\bibinfo{author}{Buchman, T.~G.}, \bibinfo{author}{Kubos, K.~L.},
  \bibinfo{author}{Seidler, A.~J.} \& \bibinfo{author}{Siegforth, M.~J.}
\newblock \bibinfo{journal}{\bibinfo{title}{A comparison of statistical and
  connectionist models for the prediction of chronicity in a surgical intensive
  care unit.}}
\newblock {\emph{\JournalTitle{Crit. Care Med.}}}
  \textbf{\bibinfo{volume}{22}}, \bibinfo{pages}{750--762}
  (\bibinfo{year}{1994}).

\bibitem{yousefi2016learning}
\bibinfo{author}{Yousefi, S.}, \bibinfo{author}{Song, C.},
  \bibinfo{author}{Nauata, N.} \& \bibinfo{author}{Cooper, L.}
\newblock \bibinfo{journal}{\bibinfo{title}{Learning genomic representations to
  predict clinical outcomes in cancer}}.
\newblock {Preprint at \url{https://arxiv.org/abs/1609.08663}}
 (\bibinfo{year}{2016}).

\bibitem{ranganath2016deep}
\bibinfo{author}{Ranganath, R.}, \bibinfo{author}{Perotte, A.},
  \bibinfo{author}{Elhadad, N.} \& \bibinfo{author}{Blei, D.}
\newblock \bibinfo{title}{Deep survival analysis}.
\newblock In \emph{\bibinfo{booktitle}{Proceedings of the 1st Machine Learning
  for Healthcare Conference}}, vol.~\bibinfo{volume}{56}
  (\bibinfo{publisher}{PMLR}, \bibinfo{address}{Los Angeles, California, USA},
  \bibinfo{year}{2016}).

\bibitem{lasko2013plosone}
\bibinfo{author}{Lasko, T.~A.}, \bibinfo{author}{Denny, J.~C.} \&
  \bibinfo{author}{Levy, M.~A.}
\newblock \bibinfo{journal}{\bibinfo{title}{Computational phenotype discovery
  using unsupervised feature learning over noisy, sparse, and irregular
  clinical data}}.
\newblock {\emph{\JournalTitle{PLoS ONE}}} \textbf{\bibinfo{volume}{8}},
  \bibinfo{pages}{e66341} (\bibinfo{year}{2013}).

\bibitem{che2015deep}
\bibinfo{author}{Che, Z.}, \bibinfo{author}{Kale, D.}, \bibinfo{author}{Li,
  W.}, \bibinfo{author}{Bahadori, M.~T.} \& \bibinfo{author}{Liu, Y.}
\newblock \bibinfo{title}{Deep computational phenotyping}.
\newblock In \emph{\bibinfo{booktitle}{Proceedings of the 21th ACM SIGKDD
  International Conference on Knowledge Discovery and Data Mining}},
  \bibinfo{pages}{507--516} (\bibinfo{publisher}{ACM},
  \bibinfo{address}{Sydney, Australia}, \bibinfo{year}{2015}).

\bibitem{choi2015doctor}
\bibinfo{author}{Choi, E.}, \bibinfo{author}{Bahadori, M.~T.},
  \bibinfo{author}{Schuetz, A.}, \bibinfo{author}{Stewart, W.~F.} \&
  \bibinfo{author}{Sun, J.}
\newblock \bibinfo{title}{Doctor {AI}: Predicting clinical events via recurrent
  neural networks}.
\newblock In \emph{\bibinfo{booktitle}{Proceedings of the 1st Machine Learning
  for Healthcare Conference}}, vol.~\bibinfo{volume}{56}
  (\bibinfo{publisher}{PMLR}, \bibinfo{address}{Los Angeles, California, USA},
  \bibinfo{year}{2016}).

\bibitem{razavian2016multi}
\bibinfo{author}{Razavian, N.}, \bibinfo{author}{Marcus, J.} \&
  \bibinfo{author}{Sontag, D.}
\newblock \bibinfo{title}{Multi-task prediction of disease onsets from
  longitudinal lab tests}.
\newblock In \emph{\bibinfo{booktitle}{Proceedings of the 1st Machine Learning
  for Healthcare Conference}}, vol.~\bibinfo{volume}{56}
  (\bibinfo{publisher}{PMLR}, \bibinfo{address}{Los Angeles, California, USA},
  \bibinfo{year}{2016}).

\bibitem{lipton2016learning}
\bibinfo{author}{Lipton, Z.~C.}, \bibinfo{author}{Kale, D.~C.},
  \bibinfo{author}{Elkan, C.} \& \bibinfo{author}{Wetzel, R.}
\newblock \bibinfo{title}{Learning to diagnose with {LSTM} recurrent neural
  networks}.
\newblock In \emph{\bibinfo{booktitle}{International Conference on Learning
  Representations}} (\bibinfo{address}{San Juan, Puerto Rico},
  \bibinfo{year}{2016}).

\bibitem{ngufor2015multi}
\bibinfo{author}{Ngufor, C.}, \bibinfo{author}{Upadhyaya, S.},
  \bibinfo{author}{Murphree, D.}, \bibinfo{author}{Kor, D.} \&
  \bibinfo{author}{Pathak, J.}
\newblock \bibinfo{title}{Multi-task learning with selective cross-task
  transfer for predicting bleeding and other important patient outcomes}.
\newblock In \emph{\bibinfo{booktitle}{IEEE International Conference on Data
  Science and Advanced Analytics (DSAA)}}, \bibinfo{pages}{1--8}
  (\bibinfo{organization}{IEEE}, \bibinfo{address}{Paris, France},
  \bibinfo{year}{2015}).

\bibitem{collobert2008unified}
\bibinfo{author}{Collobert, R.} \& \bibinfo{author}{Weston, J.}
\newblock \bibinfo{title}{A unified architecture for natural language
  processing: Deep neural networks with multitask learning}.
\newblock In \emph{\bibinfo{booktitle}{Proceedings of the 25th international
  conference on Machine learning}}, \bibinfo{pages}{160--167}
  (\bibinfo{organization}{ACM}, \bibinfo{address}{Helsinki, Finland},
  \bibinfo{year}{2008}).

\bibitem{TCS-first}
\bibinfo{author}{Gupta, P.}, \bibinfo{author}{Malhotra, P.},
  \bibinfo{author}{Vig, L.} \& \bibinfo{author}{Shroff, G.}
\newblock \bibinfo{title}{Using features from pre-trained timenet for clinical
  predictions}.
\newblock In \emph{\bibinfo{booktitle}{Proceedings of the 3rd International
  Workshop on Knowledge Discovery in Healthcare Data at IJCAI-ECAI}},
  \bibinfo{pages}{38--44} (\bibinfo{address}{Stockholm, Sweden},
  \bibinfo{year}{2018}).

\bibitem{TCS-transfer}
\bibinfo{author}{Gupta, P.}, \bibinfo{author}{Malhotra, P.},
  \bibinfo{author}{Vig, L.} \& \bibinfo{author}{Shroff, G.}
\newblock \bibinfo{title}{Transfer learning for clinical time series analysis
  using recurrent neural networks}.
\newblock In \emph{\bibinfo{booktitle}{Machine Learning for Medicine and
  Healthcare Workshop at ACM KDD 2018 Conference}} (\bibinfo{address}{London,
  United Kingdom}, \bibinfo{year}{2018}).

\bibitem{amazon-NER-2018}
\bibinfo{author}{Jin, M.} \emph{et~al.}
\newblock \bibinfo{title}{Improving hospital mortality prediction with medical
  named entities and multimodal learning}.
\newblock In \emph{\bibinfo{booktitle}{Machine Learning for Health (ML4H)
  Workshop at NeurIPS}} (\bibinfo{address}{Montreal, Canada},
  \bibinfo{year}{2018}).

\bibitem{Clinical-SequenceTransformerNetworks2018}
\bibinfo{author}{Oh, J.}, \bibinfo{author}{Wang, J.} \& \bibinfo{author}{Wiens,
  J.}
\newblock \bibinfo{title}{Learning to exploit invariances in clinical
  time-series data using sequence transformer networks}.
\newblock In \emph{\bibinfo{booktitle}{Proceedings of the 3rd Machine Learning
  for Healthcare Conference}}, vol.~\bibinfo{volume}{85},
  \bibinfo{pages}{332--347} (\bibinfo{publisher}{PMLR}, \bibinfo{address}{Palo
  Alto, California, USA}, \bibinfo{year}{2018}).

\bibitem{MissingDataRepresentations2019}
\bibinfo{author}{Malone, B.}, \bibinfo{author}{Garcia-Duran, A.} \&
  \bibinfo{author}{Niepert, M.}
\newblock \bibinfo{journal}{\bibinfo{title}{Learning representations of missing
  data for predicting patient outcomes}}.
\newblock {Preprint at \url{https://arxiv.org/abs/1811.04752}}
 (\bibinfo{year}{2018}).

\bibitem{AdverseEventForecastingRL2018}
\bibinfo{author}{Chang, C.-H.}, \bibinfo{author}{Mai, M.} \&
  \bibinfo{author}{Goldenberg, A.}
\newblock \bibinfo{title}{Dynamic measurement scheduling for adverse event
  forecasting using deep RL}.
\newblock In \emph{\bibinfo{booktitle}{Machine Learning for Health (ML4H)
  Workshop at NeurIPS}} (\bibinfo{address}{Montreal, Canada},
  \bibinfo{year}{2018}).

\bibitem{RAIM2018}
\bibinfo{author}{Xu, Y.}, \bibinfo{author}{Biswal, S.},
  \bibinfo{author}{Deshpande, S.~R.}, \bibinfo{author}{Maher, K.~O.} \&
  \bibinfo{author}{Sun, J.}
\newblock \bibinfo{title}{Raim: Recurrent attentive and intensive model of
  multimodal patient monitoring data}.
\newblock In \emph{\bibinfo{booktitle}{Proceedings of the 24th ACM SIGKDD
  International Conference on Knowledge Discovery \& Data Mining}},
  \bibinfo{pages}{2565--2573} (\bibinfo{organization}{ACM},
  \bibinfo{address}{London, United Kingdom}, \bibinfo{year}{2018}).

\bibitem{RNN-GP2018chung}
\bibinfo{author}{Chung, I.}, \bibinfo{author}{Kim, S.}, \bibinfo{author}{Lee,
  J.}, \bibinfo{author}{Hwang, S.~J.} \& \bibinfo{author}{Yang, E.}
\newblock \bibinfo{journal}{\bibinfo{title}{Mixed effect composite RNN-GP: A personalized and reliable prediction model for healthcare}}.
\newblock {Preprint at \url{https://arxiv.org/abs/1806.01551}}
 (\bibinfo{year}{2018}).
  

\bibitem{taha}
\bibinfo{author}{Bahadori, M.~T.}
\newblock \bibinfo{title}{Spectral capsule networks}.
\newblock In \emph{\bibinfo{booktitle}{International Conference on Learning
  Representations Workshop Track}} (\bibinfo{address}{New Orleans, Louisiana,
  USA}, \bibinfo{year}{2018}).

\bibitem{TAMUreadmissions}
\bibinfo{author}{Rafi, P.}, \bibinfo{author}{Pakbin, A.} \&
  \bibinfo{author}{Pentyala, S.~K.}
\newblock \bibinfo{title}{Interpretable deep learning framework for predicting
  all-cause 30-day ICU readmissions}.
\newblock \bibinfo{type}{\textit{Tech. Rep.}}, (\bibinfo{institution}{{Texas A\&M
  University}}, \bibinfo{year}{2018}).

\bibitem{attendanddiagnose}
\bibinfo{author}{Song, H.}, \bibinfo{author}{Rajan, D.},
  \bibinfo{author}{Thiagarajan, J.~J.} \& \bibinfo{author}{Spanias, A.}
\newblock \bibinfo{title}{Attend and diagnose: Clinical time series analysis
  using attention models}.
\newblock In \emph{\bibinfo{booktitle}{Proceedings of the Thirty-Second {AAAI}
  Conference on Artificial Intelligence}} (\bibinfo{publisher}{AAAI Press},
  \bibinfo{address}{New Orleans, Louisiana, USA}, \bibinfo{year}{2018}).

\bibitem{purushotham2017benchmark}
\bibinfo{author}{Purushotham, S.}, \bibinfo{author}{Meng, C.},
  \bibinfo{author}{Che, Z.} \& \bibinfo{author}{Liu, Y.}
\newblock \bibinfo{journal}{\bibinfo{title}{Benchmarking deep learning models
  on large healthcare datasets}}.
\newblock {\emph{\JournalTitle{J. Biomed. Inform.}}}
  \textbf{\bibinfo{volume}{83}}, \bibinfo{pages}{112 -- 134}
  (\bibinfo{year}{2018}).

\bibitem{le1984simplified}
\bibinfo{author}{Le~Gall, J.-R.} \emph{et~al.}
\newblock \bibinfo{journal}{\bibinfo{title}{A simplified acute physiology score
  for icu patients.}}
\newblock {\emph{\JournalTitle{Crit. Care Med.}}}
  \textbf{\bibinfo{volume}{12}}, \bibinfo{pages}{975--977}
  (\bibinfo{year}{1984}).

\bibitem{silva2012physionet}
\bibinfo{author}{Silva, I.}, \bibinfo{author}{Moody, G.},
  \bibinfo{author}{Scott, D.~J.}, \bibinfo{author}{Celi, L.~A.} \&
  \bibinfo{author}{Mark, R.~G.}
\newblock \bibinfo{title}{Predicting in-hospital mortality of icu patients: The
  physionet/computing in cardiology challenge 2012}.
\newblock In \emph{\bibinfo{booktitle}{2012 Computing in Cardiology}},
  \bibinfo{pages}{245--248} (\bibinfo{publisher}{IEEE},
  \bibinfo{address}{Krakow, Poland}, \bibinfo{year}{2012}).

\bibitem{marlin2012unsupervised}
\bibinfo{author}{Marlin, B.~M.}, \bibinfo{author}{Kale, D.~C.},
  \bibinfo{author}{Khemani, R.~G.} \& \bibinfo{author}{Wetzel, R.~C.}
\newblock \bibinfo{title}{Unsupervised pattern discovery in electronic health
  care data using probabilistic clustering models}.
\newblock In \emph{\bibinfo{booktitle}{Proceedings of the 2nd ACM SIGHIT
  International Health Informatics Symposium}}, \bibinfo{pages}{389--398}
  (\bibinfo{organization}{ACM}, \bibinfo{address}{Miami, Florida},
  \bibinfo{year}{2012}).

\bibitem{Davis2006-roc-pr}
\bibinfo{author}{Davis, J.} \& \bibinfo{author}{Goadrich, M.}
\newblock \bibinfo{title}{The relationship between precision-recall and roc
  curves}.
\newblock In \emph{\bibinfo{booktitle}{Proceedings of the 23rd International
  Conference on Machine Learning}}, \bibinfo{pages}{233--240}
  (\bibinfo{publisher}{ACM}, \bibinfo{address}{Pittsburgh, Pennsylvania, USA},
  \bibinfo{year}{2006}).

\bibitem{subbe2001validation}
\bibinfo{author}{Subbe, C.}, \bibinfo{author}{Kruger, M.},
  \bibinfo{author}{Rutherford, P.} \& \bibinfo{author}{Gemmel, L.}
\newblock \bibinfo{journal}{\bibinfo{title}{Validation of a modified early
  warning score in medical admissions}}.
\newblock {\emph{\JournalTitle{Qjm}}} \textbf{\bibinfo{volume}{94}},
  \bibinfo{pages}{521--526} (\bibinfo{year}{2001}).

\bibitem{prytherch2010views}
\bibinfo{author}{Prytherch, D.~R.}, \bibinfo{author}{Smith, G.~B.},
  \bibinfo{author}{Schmidt, P.~E.} \& \bibinfo{author}{Featherstone, P.~I.}
\newblock \bibinfo{journal}{\bibinfo{title}{Views -- towards a national early   warning score for detecting adult inpatient deterioration}}.
\newblock {\emph{\JournalTitle{Resuscitation}}} \textbf{\bibinfo{volume}{81}},
  \bibinfo{pages}{932--937} (\bibinfo{year}{2010}).

\bibitem{aleks2009probabilistic}
\bibinfo{author}{Aleks, N.} \emph{et~al.}
\newblock \bibinfo{title}{Probabilistic detection of short events, with
  application to critical care monitoring}.
\newblock In \emph{\bibinfo{booktitle}{Advances in Neural Information
  Processing Systems 21}}, \bibinfo{pages}{49--56} (\bibinfo{publisher}{Curran
  Associates, Inc.}, \bibinfo{address}{Vancouver, Canada},
  \bibinfo{year}{2009}).

\bibitem{henry2015targeted}
\bibinfo{author}{Henry, K.~E.}, \bibinfo{author}{Hager, D.~N.},
  \bibinfo{author}{Pronovost, P.~J.} \& \bibinfo{author}{Saria, S.}
\newblock \bibinfo{journal}{\bibinfo{title}{A targeted real-time early warning score (trewscore) for septic shock}}.
\newblock {\emph{\JournalTitle{Sci. Transl. Med.}}}
  \textbf{\bibinfo{volume}{7}}, \bibinfo{pages}{299ra122--299ra122}
  (\bibinfo{year}{2015}).

\bibitem{clifton2012gaussian}
\bibinfo{author}{Clifton, L.}, \bibinfo{author}{Clifton, D.~A.},
  \bibinfo{author}{Pimentel, M.~A.}, \bibinfo{author}{Watkinson, P.~J.} \&
  \bibinfo{author}{Tarassenko, L.}
\newblock \bibinfo{title}{Gaussian process regression in vital-sign early
  warning systems}.
\newblock In \emph{\bibinfo{booktitle}{2012 Annual International Conference of
  the IEEE Engineering in Medicine and Biology Society}},
  \bibinfo{pages}{6161--6164} (\bibinfo{organization}{IEEE},
  \bibinfo{address}{San Diego, California, USA}, \bibinfo{year}{2012}).

\bibitem{AHRQ}
\bibinfo{author}{Romano, P.}, \bibinfo{author}{Hussey P.} \& \bibinfo{author}{Ritley, D.}.
\newblock \bibinfo{title}{\textit{Selecting quality and resource use measures: A decision guide for community quality collaboratives}}. (Agency for Healthcare Research and Quality, \bibinfo{year}{2014}).

\bibitem{higgins2003early}
\bibinfo{author}{Higgins, T.~L.} \emph{et~al.}
\newblock \bibinfo{journal}{\bibinfo{title}{Early indicators of prolonged
  intensive care unit stay: Impact of illness severity, physician staffing, and
  pre--intensive care unit length of stay}}.
\newblock {\emph{\JournalTitle{Crit. Care Med.}}}
  \textbf{\bibinfo{volume}{31}}, \bibinfo{pages}{45--51}
  (\bibinfo{year}{2003}).

\bibitem{osler1998predicting}
\bibinfo{author}{Osler, T.~M.} \emph{et~al.}
\newblock \bibinfo{journal}{\bibinfo{title}{Predicting survival, length of
  stay, and cost in the surgical intensive care unit: Apache ii versus iciss}}.
\newblock {\emph{\JournalTitle{Journal of Trauma and Acute Care Surgery}}}
  \textbf{\bibinfo{volume}{45}}, \bibinfo{pages}{234--238}
  (\bibinfo{year}{1998}).

\bibitem{paterson2006prediction}
\bibinfo{author}{Paterson, R.} \emph{et~al.}
\newblock \bibinfo{journal}{\bibinfo{title}{Prediction of in-hospital mortality
  and length of stay using an early warning scoring system: clinical audit}}.
\newblock {\emph{\JournalTitle{Clinical Medicine}}}
  \textbf{\bibinfo{volume}{6}}, \bibinfo{pages}{281--284}
  (\bibinfo{year}{2006}).

\bibitem{pofahl1998use}
\bibinfo{author}{Pofahl, W.~E.}, \bibinfo{author}{Walczak, S.~M.},
  \bibinfo{author}{Rhone, E.} \& \bibinfo{author}{Izenberg, S.~D.}
\newblock \bibinfo{journal}{\bibinfo{title}{Use of an artificial neural network
  to predict length of stay in acute pancreatitis}}.
\newblock {\emph{\JournalTitle{The American Surgeon}}}
  \textbf{\bibinfo{volume}{64}}, \bibinfo{pages}{868} (\bibinfo{year}{1998}).

\bibitem{cohen1960coefficient}
\bibinfo{author}{Cohen, J.}
\newblock \bibinfo{journal}{\bibinfo{title}{A coefficient of agreement for
  nominal scales}}.
\newblock {\emph{\JournalTitle{Educational and Psychological Measurement}}}
  \textbf{\bibinfo{volume}{20}}, \bibinfo{pages}{37--46}
  (\bibinfo{year}{1960}).

\bibitem{Cohen-uses-brennan1981coefficient}
\bibinfo{author}{Brennan, R.~L.} \& \bibinfo{author}{Prediger, D.~J.}
\newblock \bibinfo{journal}{\bibinfo{title}{Coefficient kappa: Some uses, misuses, and alternatives}}.
\newblock {\emph{\JournalTitle{Educ. Psychol. Meas.}}}
  \textbf{\bibinfo{volume}{41}}, \bibinfo{pages}{687--699}
  (\bibinfo{year}{1981}).

\bibitem{oellrich2015digital}
\bibinfo{author}{Oellrich, A.} \emph{et~al.}
\newblock \bibinfo{journal}{\bibinfo{title}{{The digital revolution in phenotyping}}}.
\newblock {\emph{\JournalTitle{Brief. Bioinform.}}}
  \textbf{\bibinfo{volume}{17}}, \bibinfo{pages}{819--830}
  (\bibinfo{year}{2015}).

\bibitem{agarwal2016noisy}
\bibinfo{author}{Agarwal, V.} \emph{et~al.}
\newblock \bibinfo{journal}{\bibinfo{title}{Learning statistical models of
  phenotypes using noisy labeled training data}}.
\newblock {\emph{\JournalTitle{Journal of the American Medical Informatics
  Association}}} \textbf{\bibinfo{volume}{23}}, \bibinfo{pages}{1166}
  (\bibinfo{year}{2016}).

\bibitem{halpern2016anchors}
\bibinfo{author}{Halpern, Y.}, \bibinfo{author}{Horng, S.},
  \bibinfo{author}{Choi, Y.} \& \bibinfo{author}{Sontag, D.}
\newblock \bibinfo{journal}{\bibinfo{title}{Electronic medical record
  phenotyping using the anchor and learn framework}}.
\newblock {\emph{\JournalTitle{Journal of the American Medical Informatics
  Association}}} \textbf{\bibinfo{volume}{23}}, \bibinfo{pages}{731}
  (\bibinfo{year}{2016}).

\bibitem{ho2014mhp}
\bibinfo{author}{Ho, J.~C.}, \bibinfo{author}{Ghosh, J.} \&
  \bibinfo{author}{Sun, J.}
\newblock \bibinfo{title}{Marble: High-throughput phenotyping from electronic
  health records via sparse nonnegative tensor factorization}.
\newblock In \emph{\bibinfo{booktitle}{Proceedings of the 20th ACM SIGKDD
  International Conference on Knowledge Discovery and Data Mining}},
  \bibinfo{pages}{115--124} (\bibinfo{publisher}{ACM}, \bibinfo{address}{New
  York, New York, USA}, \bibinfo{year}{2014}).

\bibitem{CCS}
\newblock \bibinfo{title}{\textit{Clinical Classifications Software (CCS) for ICD-9-CM fact sheet}} (Agency for Healthcare Research and Quality, \bibinfo{year}{2012}).

\bibitem{LSTM-hochreiter1997}
\bibinfo{author}{Hochreiter, S.} \& \bibinfo{author}{Schmidhuber, J.}
\newblock \bibinfo{journal}{\bibinfo{title}{Long short-term memory}}.
\newblock {\emph{\JournalTitle{Neur. Comp.}}}
  \textbf{\bibinfo{volume}{9}}, \bibinfo{pages}{1735--1780}
  (\bibinfo{year}{1997}).

\bibitem{peephole-gers2000recurrent}
\bibinfo{author}{{Gers}, F.~A.} \& \bibinfo{author}{{Schmidhuber}, J.}
\newblock \bibinfo{title}{Recurrent nets that time and count}.
\newblock In \emph{\bibinfo{booktitle}{Proceedings of the IEEE-INNS-ENNS
  International Joint Conference on Neural Networks}},
  vol.~\bibinfo{volume}{3}, \bibinfo{pages}{189--194}
  (\bibinfo{organization}{IEEE}, \bibinfo{address}{Como, Italy},
  \bibinfo{year}{2000}).

\bibitem{lipton2016missing}
\bibinfo{author}{Lipton, Z.~C.}, \bibinfo{author}{Kale, D.~C.} \&
  \bibinfo{author}{Wetzel, R.}
\newblock \bibinfo{title}{Modeling missing data in clinical time series with
  rnns}.
\newblock In \emph{\bibinfo{booktitle}{Proceedings of the 1st Machine Learning
  for Healthcare Conference}}, vol.~\bibinfo{volume}{56}
  (\bibinfo{publisher}{PMLR}, \bibinfo{address}{Los Angeles, California, USA},
  \bibinfo{year}{2016}).

\bibitem{kingma2014adam}
\bibinfo{author}{Kingma, D.} \& \bibinfo{author}{Ba, J.}
\newblock \bibinfo{journal}{\bibinfo{title}{Adam: A method for stochastic optimization}}.
\newblock {Preprint at \url{https://arxiv.org/abs/1412.6980}}
 (\bibinfo{year}{2014}).

\bibitem{choi2016retain}
\bibinfo{author}{Choi, E.} \emph{et~al.}
\newblock \bibinfo{title}{Retain: An interpretable predictive model for
  healthcare using reverse time attention mechanism}.
\newblock In \emph{\bibinfo{booktitle}{Advances in Neural Information
  Processing Systems 29}}, \bibinfo{pages}{3504--3512}
  (\bibinfo{publisher}{Curran Associates, Inc.}, \bibinfo{address}{Barcelona,
  Spain}, \bibinfo{year}{2016}).

\bibitem{BioCreativeII}
\bibinfo{author}{Smith, L.} \emph{et~al.}
\newblock \bibinfo{journal}{\bibinfo{title}{Overview of biocreative ii gene
  mention recognition}}.
\newblock {\emph{\JournalTitle{Genome Biology}}} \textbf{\bibinfo{volume}{9}},
  \bibinfo{pages}{S2} (\bibinfo{year}{2008}).

\bibitem{CheXNet}
\bibinfo{author}{Rajpurkar, P.} \emph{et~al.}
\newblock \bibinfo{journal}{\bibinfo{title}{CheXNet: Radiologist-level pneumonia detection on chest X-rays with deep learning}}.
\newblock {Preprint at \url{https://arxiv.org/abs/1711.05225}}
 (\bibinfo{year}{2017}).

\bibitem{mimicbenchmarkrepo}
\bibinfo{author}{Harutyunyan, H.} \emph{et~al.}
\bibinfo{journal}{\bibinfo{title}{MIMIC-III benchmark repository}}.
\newblock {\emph{\JournalTitle{Zenodo}}}
  \url{https://doi.org/10.5281/zenodo.1306527}
  (\bibinfo{year}{2018}).

\end{thebibliography}

\end{document}